%% file: main.tex
\documentclass[final]{cvpr}

\usepackage{times}
\usepackage{epsfig}
\usepackage{graphicx}
\usepackage{amsmath}
\usepackage{amssymb}
\usepackage{algorithm}
\usepackage{booktabs}
\usepackage{microtype}
\usepackage{algpseudocode}
\usepackage{multirow}
\usepackage{float}
\usepackage[toc,page]{appendix}
\usepackage[caption = false]{subfig}
\usepackage{flushend}

\usepackage{caption}

\usepackage[pagebackref=true,breaklinks=true,colorlinks,bookmarks=false]{hyperref}
\usepackage{cleveref}
\def\cvprPaperID{6375} 
\def\confYear{CVPR 2021}
\input{defs}

\title{Invisible Perturbations:
  Physical Adversarial Examples \\Exploiting the Rolling Shutter Effect}

\author{Athena Sayles\thanks{Both authors contributed equally to this work.},\, Ashish Hooda\footnote[1]{},\, Mohit Gupta,\, Rahul Chatterjee,\, and Earlence Fernandes\\[3pt]
University of Wisconsin--Madison\\
{\tt\small \{esayles, hooda, mohitg, chatterjee, earlence\}@cs.wisc.edu}
}

\usepackage{fancyhdr}
\begin{document}


\maketitle
\thispagestyle{fancy}
\input{abstract}
\input{intro}
\input{related}

\input{bg}
\input{attack-theory}
\input{attack-physical}
\input{exp}  
\input{discussion}

\input{conclusion}
\newpage
{\small
\bibliographystyle{ieee_fullname}
\bibliography{bib}
}
\newpage
\input{appendix}
\end{document}

%% file: defs.tex
\usepackage{xspace}
\usepackage{xparse}
\usepackage{booktabs}
\usepackage{mathtools}
\usepackage{amsthm}
\usepackage{xcolor}
\DeclareMathAlphabet{\mathcal}{OMS}{cmsy}{m}{n}
\SetMathAlphabet{\mathcal}{bold}{OMS}{cmsy}{b}{n}
\newcommand{\bnm}{\begin{newmath}}
\newcommand{\enm}{\end{newmath}}

\newcommand{\bea}{\begin{neweqnarrays}}%
\newcommand{\eea}{\end{neweqnarrays}}%

\newcommand{\bne}{\begin{newequation}}
\newcommand{\ene}{\end{newequation}}

\newcommand{\bal}{\begin{newalign}}
\newcommand{\eal}{\end{newalign}}

\newenvironment{newalign}{\begin{align*}%
\setlength{\abovedisplayskip}{4pt}%
\setlength{\belowdisplayskip}{4pt}%
\setlength{\abovedisplayshortskip}{6pt}%
\setlength{\belowdisplayshortskip}{6pt} }{\end{align*}}

\newenvironment{newmath}{\begin{displaymath}%
\setlength{\abovedisplayskip}{4pt}%
\setlength{\belowdisplayskip}{4pt}%
\setlength{\abovedisplayshortskip}{6pt}%
\setlength{\belowdisplayshortskip}{6pt} }{\end{displaymath}}

\newenvironment{neweqnarrays}{\begin{eqnarray*}%
\setlength{\abovedisplayskip}{-1pt}%
\setlength{\belowdisplayskip}{-1pt}%
\setlength{\abovedisplayshortskip}{1pt}%
\setlength{\belowdisplayshortskip}{1pt}%
\setlength{\jot}{-0.4in} }{\end{eqnarray*}}

\newenvironment{newequation}{\begin{equation}%
\setlength{\abovedisplayskip}{4pt}%
\setlength{\belowdisplayskip}{4pt}%
\setlength{\abovedisplayshortskip}{6pt}%
\setlength{\belowdisplayshortskip}{6pt} }{\end{equation}}

\newcounter{ctr}

\newenvironment{newitemize}{%
\begin{list}{\mbox{}\hspace{5pt}$\bullet$\hfill}{\labelwidth=15pt%
\labelsep=4pt \leftmargin=12pt \topsep=3pt%
\setlength{\listparindent}{\saveparindent}%
\setlength{\parsep}{\saveparskip}%
\setlength{\itemsep}{3pt} }}{\end{list}}

\newlength{\saveparindent}
\newlength{\saveparskip}
\DeclareMathSymbol{\mlq}{\mathord}{operators}{``}
\DeclareMathSymbol{\mrq}{\mathord}{operators}{`'}

\newcommand{\E}{{\rm I\kern-.3em E}}

\newcommand{\getsr}{{\:{\leftarrow{\hspace*{-3pt}\raisebox{.75pt}{$\scriptscriptstyle\$$}}}\:}}

\renewcommand{\algref}[1]{\mbox{Algorithm~\ref{#1}}}
\renewcommand{\eqref}[1]{Eq.~\mbox{(\ref{#1})}}

\renewcommand{\tabref}[1]{\mbox{Table~\ref{#1}}}

\newcommand{\tabfontsize}{\scriptsize}
\newcommand{\gamesfontsize}{\footnotesize}

\providecommand{\setstretch}[1]{}

\newcommand{\Z}{\mathbb{Z}}

\newcommand{\thh}{^{\textit{th}}} 

\def \part {part}

\renewcommand{\paragraph}[1]{\vspace*{6pt}\noindent\textbf{#1}\;}

\newcounter{mytable}
\def\mytable{\begin{centering}\refstepcounter{mytable}}
\def\endmytable{\end{centering}}

\newcounter{myfig}
\def\myfig{\begin{centering}\refstepcounter{myfig}}
\def\endmyfig{\end{centering}}

\def \blackslug{\hbox{\hskip 1pt \vrule width 4pt height 8pt
    depth 1.5pt \hskip 1pt}}
\def \qed{\quad\blackslug\lower 8.5pt\null\par}

\newcounter{mynote}[section]

\newcommand\ignore[1]{}

\newcounter{rcnote}[section]

\newcounter{ahnote}[section]

\DeclareMathOperator{\Exp}{\mathbb{E}}

\newcounter{asnote}[section]

\newcommand{\rhf}[2]{R_{f, \gamma}}

\newcommand{\M}{{\mathcal{M}}}

\DeclareDocumentCommand{\edist}{o o}{
  \ensuremath{
    \IfNoValueTF{#1}{{d}}{{\sf d}(#1,#2)}
  }
}

\DeclareMathSymbol{\mlq}{\mathord}{operators}{``}
\DeclareMathSymbol{\mrq}{\mathord}{operators}{`'}

\newcommand{\gain}{\rho}

\newcommand{\ltex}{l_{tex}}
\newcommand{\period}{\tau}
\newcommand{\exptime}{t_e}
\newcommand{\readouttime}{t_r}
\newcommand{\signal}{f}
\newcommand{\digsignal}{\hat{f}}
\newcommand{\finalimg}{I_\mathsf{fin}} %
\newcommand{\fullimg}{I_\mathsf{full}} %
\newcommand{\fullimgn}{I_{\mathsf{full},n}} %
\newcommand{\ambimg}{I_\mathsf{amb}}
\newcommand{\sigimg}{I_\mathsf{sig}}
\newcommand{\ambimgn}{I_{\mathsf{amb}, n}}
\newcommand{\sigimgn}{I_{\mathsf{sig}, n}}
\newcommand{\targetclass}{k}
\newcommand{\shift}{\mathsf{shift}}
\newcommand{\model}{\M}
\newcommand{\colorcorr}{\mathsf{C}}
\newcommand{\viewtransform}{\mathsf{T}}
\newcommand{\foolingrate}{$84\%$}
\newcommand{\mus}{\mu s}

%% file: abstract.tex
\begin{abstract}

Physical adversarial examples for camera-based computer vision have so far been achieved through visible artifacts --- a sticker on a Stop sign, colorful borders around eyeglasses or a 3D printed object with a colorful texture. An implicit assumption here is that the perturbations must be visible so that a camera can sense them. By contrast, we contribute a procedure to generate, for the first time, physical adversarial examples that are invisible to human eyes. Rather than modifying the victim object with visible artifacts, we modify light that illuminates the object. We demonstrate how an attacker can craft a modulated light signal that adversarially illuminates a scene and causes targeted misclassifications on a state-of-the-art ImageNet deep learning model. Concretely, we exploit the radiometric rolling shutter effect in commodity cameras to create precise striping patterns that appear on images. To human eyes, it appears like the object is illuminated, but the camera creates an image with stripes that will cause ML models to output the attacker-desired classification. We conduct a range of simulation and physical experiments with LEDs, demonstrating targeted attack rates up to \foolingrate.

\end{abstract}

%% file: intro.tex
\section{Introduction}\label{sec:intro}

Recent work has established that deep learning models are susceptible to adversarial examples --- manipulations to model inputs that are inconspicuous to humans but induce the models to produce attacker-desired outputs~\cite{szegedy2014intriguing,goodfellow2014explaining,carlini2017towards}. Early work in this space investigated \emph{digital} adversarial examples where the attacker can manipulate the input vector, such as modifying pixel values directly in an image classification task. As deep learning has found increasing application in real-world systems like self-driving cars~\cite{lillicrap2015continuous,geiger2012we,openpilot}, UAVs~\cite{bou2010controller,mostegel2016uav}, and robots~\cite{zhang2015towards}, the computer vision community has made great progress in understanding \emph{physical} adversarial examples~\cite{roadsigns17,eot,sharif2016accessorize,cam-sticker,patch-attack} because this attack modality is the most realistic in physical systems. 

Existing physical attacks 
include adding stickers on Stop signs that make models output Speed limit instead~\cite{roadsigns17}, colorful patterns on eyeglass frames to trick face recognition~\cite{sharif2016accessorize}, and 3D-printed objects with specific textures~\cite{athalye2017synthesizing}. However, all existing works add artifacts to the object (such as sticker or color patterns) that are visible to a human. 
In this work, we 
generate adversarial perturbations on real-world objects that are invisible to human eyes, yet produce misclassifications. Our approach exploits the differences between human and machine vision to hide adversarial patterns.

\fboxsep=0mm
\fboxrule=1.5pt

\begin{figure}[t]
  \centering\gamesfontsize
  
  
  \setlength{\fboxrule}{2pt}
  {\includegraphics[height=1.9cm]{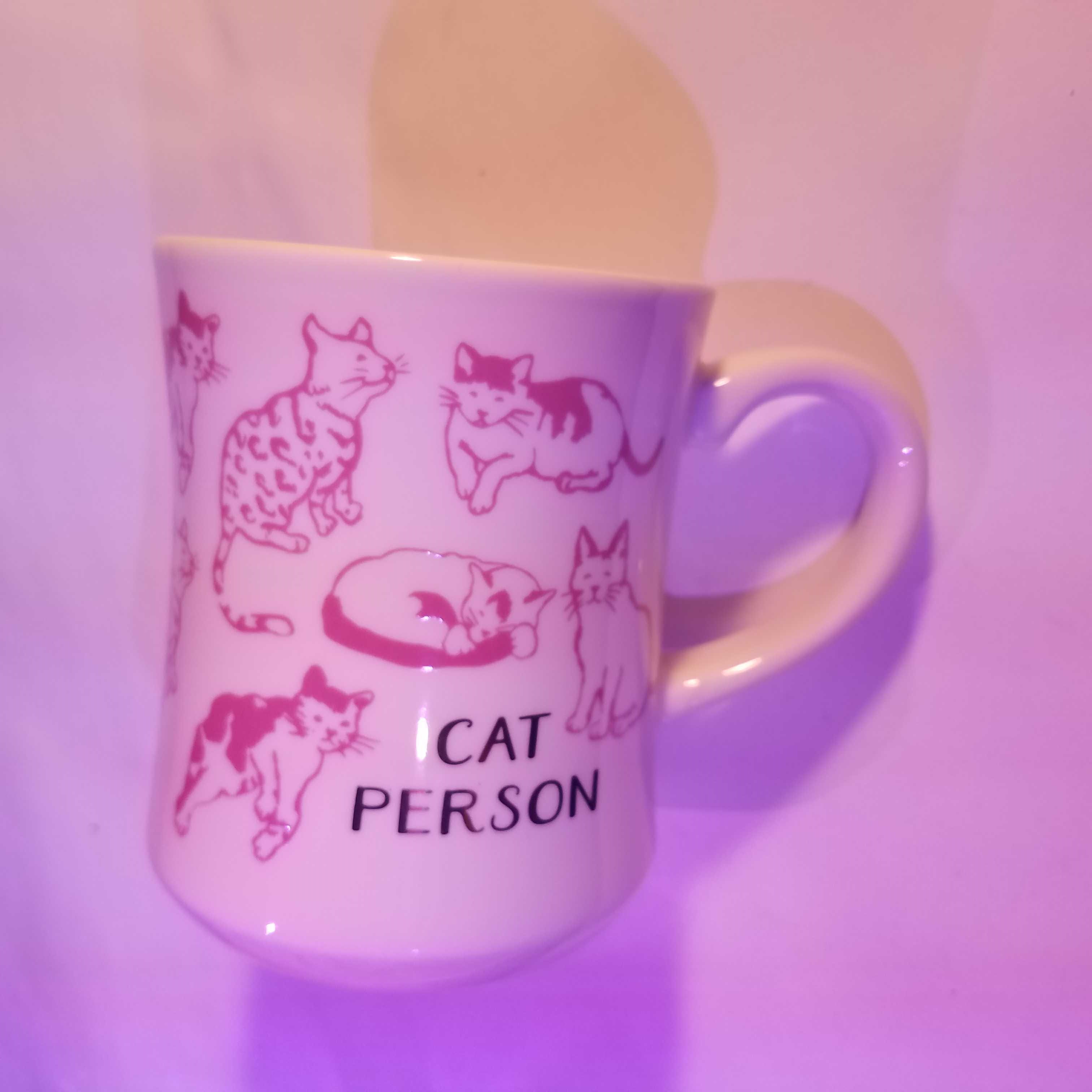}} 
  \fcolorbox{black}{gray!50}{\includegraphics[height=1.9cm]{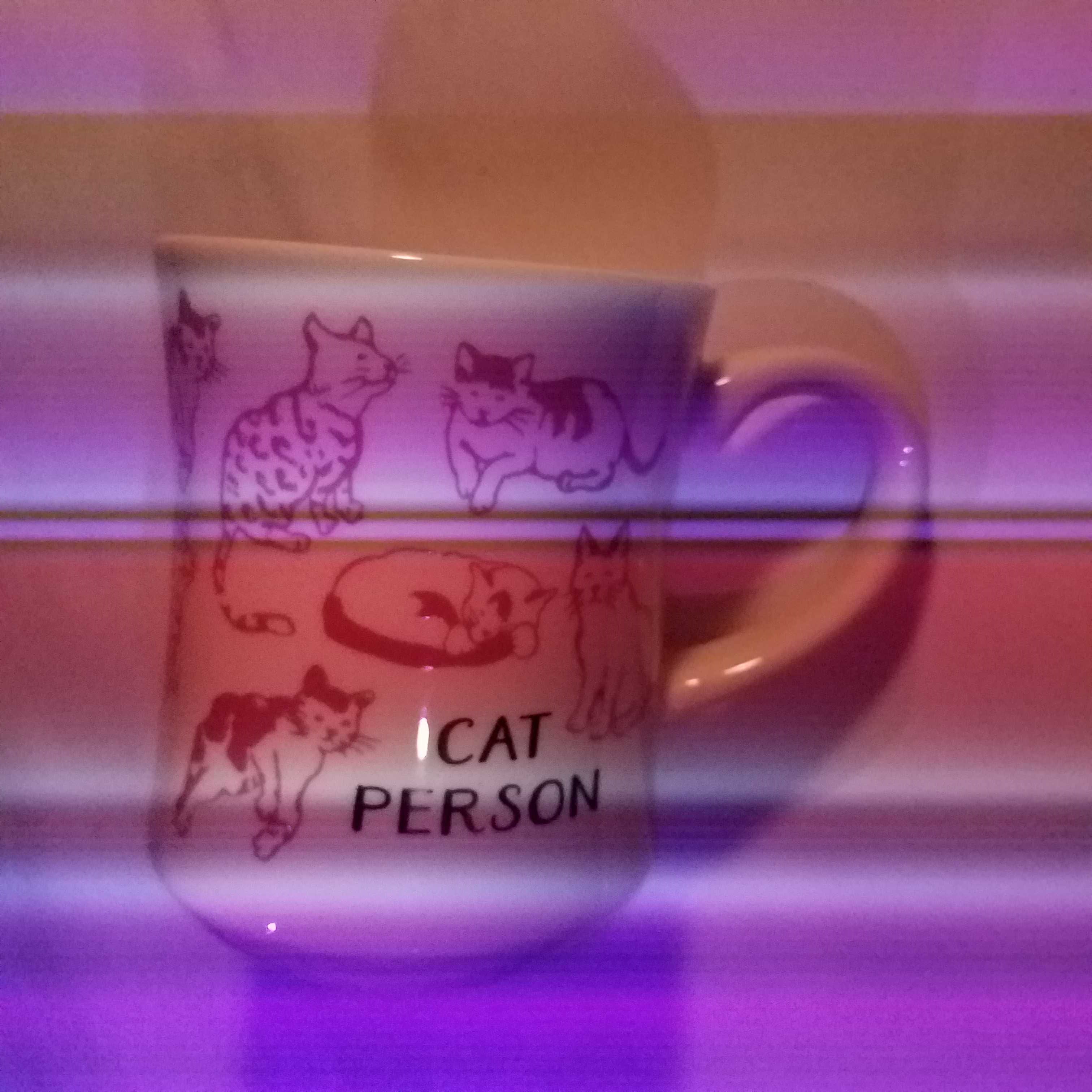}}  
  \hfill
  {\includegraphics[height=1.9cm]{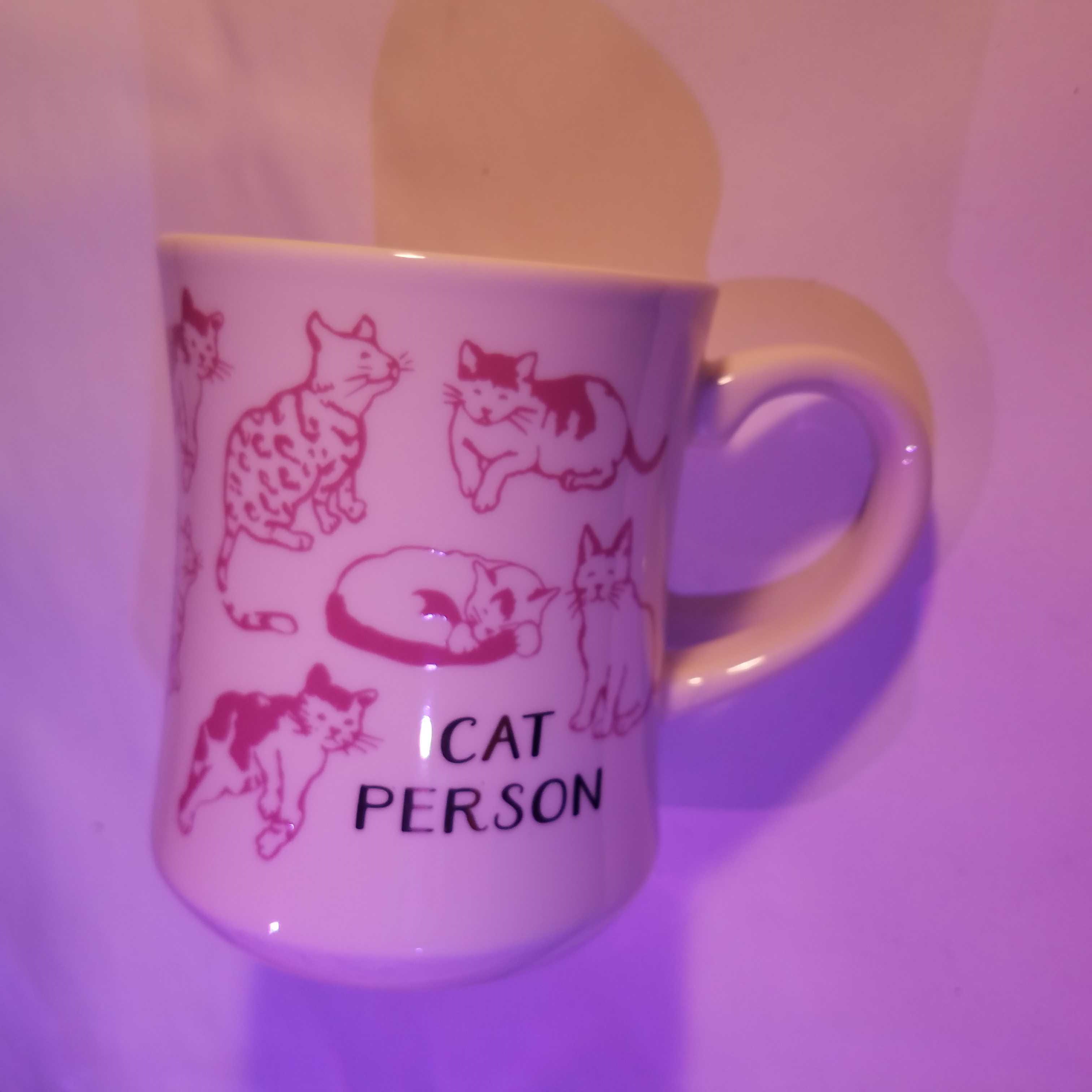}}  
  \fcolorbox{black}{gray!50}{\includegraphics[height=1.9cm]{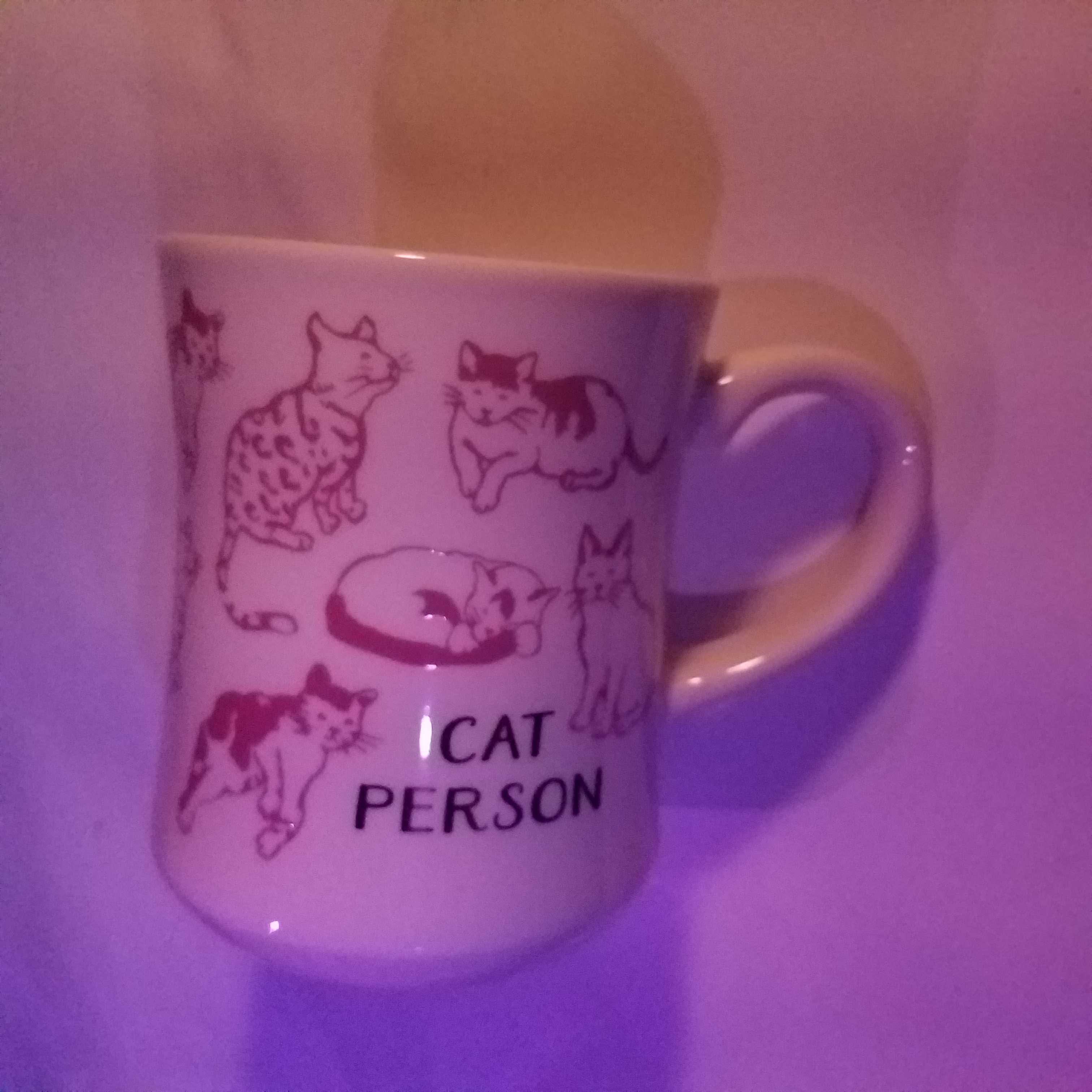}}\\ 
  \hspace{0.6cm}With Attack Signal\hspace{2.0cm} Without Attack Signal
  \caption{Images as seen by human (without border) and as captured by camera (in black border) with the attack signal (left two images) and without (right two images). 
    The image without the attack signal is classified as coffee mug (confidence $55\%$), while the image with the attack signal is classified as perfume (confidence $70\%$). The attack is robust to camera orientation, distance, and ambient lighting.}
  \vspace{-10pt}
  \label{fig:teaser}  
\end{figure}

We show an \emph{invisible physical adversarial example} in~\figref{fig:teaser}, generated by 
manipulating the light that shines on the object. The light creates adversarial patterns in the image that \emph{only} a camera perceives. In particular, we show how an attacker can exploit the \emph{radiometric} rolling shutter (RS) effect, a phenomenon that exists in rolling shutter cameras that perceive a scene whose illumination changes at a high frequency.  Digital cameras use the rolling shutter technique to obtain high resolution images at higher rate and at a cheaper price~\cite{albl2020rolling,rswhitepaper}. Rolling shutter technology is used in a majority of consumer-grade cameras, such as cellphones~\cite{kim2020object}, AR glasses~\cite{hololens} and machine vision~\cite{alliedvision,onsemi}.

%

Due to the rolling shutter effect, the adversarially-illuminated object results in an image that
contains multi-colored stripes. We contribute an algorithm 
for creating a time-varying high-frequency light pattern that can create such stripes. To the best of our knowledge, this is the first demonstration of
physical adversarial examples that exploit the radiometric rolling shutter
effect, and thus, contributes to our evolving understanding of physical attacks
on deep learning camera-based computer vision.

Similar to prior work on physical attacks, the main challenge is obtaining robustness to dynamic environmental conditions such as viewpoint and lighting. However, in our setting, there are additional environmental conditions that pose challenges in creating these attacks. Specifically: (1) Camera exposure settings influence how much of the rolling shutter effect is present, which affects the attacker's ability to craft adversarial examples.  --- long exposures lead to less pronounced rolling shutter, providing less control.
(2) The attacker's light signal can be de-synchronized with respect to the camera shutter, thus 
causing the camera to capture the adversarial signal at different offsets
causing the striping pattern to appear at different locations on the image, that can destroy its adversarial property.
(3) The space of possible perturbations is limited compared to existing attacks. Unlike sticker attacks or 3D objects that can change the victim object's texture, our attack only permits striped patterns that contain a limited set of translucent colors.  
(4) Difference in the light produced by RGB LEDs and the color perceived by camera sensor makes it harder to realize a physical signal. 

To tackle the above challenges, we create a simulation framework that captures these environmental and camera imaging conditions. The simulation is based on a differentiable analytical model of image formation and light signal transmission and reception when the \emph{radiometric} rolling shutter effect is present. Using the analytical model, we then formulate an optimization objective that we can solve using standard gradient-based methods to compute an adversarial light signal that is robust to these unique environmental and camera imaging conditions. We fabricate this light signal using programmable LEDs.

Although light-based adversarial examples are limited in the types of perturbation patterns compared to sticker-based ones, they have several advantages: (1) The attack is stealthier than sticker-based ones, as the attacker can simply turn the light source to a constant value to turn OFF the attack. (2) Unlike prior work using sticker or 3D printed object, the perturbation is not visible to human eyes. (3) The attack is dynamic and can change on-the-fly --- in a sticker-based attack, once the sticker has been placed, the attack effect cannot be changed unless the sticker is physically replaced. In our setting, the attacker can simply change the light signal and thus, change the adversarial effect. 

We characterize this new style of invisible physical adversarial example using a state-of-the-art ResNet-101 classifier trained using ImageNet~\cite{deng2009imagenet}. We conduct physical testing of our attack algorithm under various viewpoints, ambient lighting conditions, and camera exposure settings. For example, for the coffee mug shown in~\figref{fig:teaser} we obtain a targeted fooling rate of \foolingrate under a variety of conditions. We find that the attack success rate is dependent on the camera exposure setting: exposure rates shorter than $1/750s$ produce the most successful and robust attacks. 


\noindent The main contributions of our work are the following:
\begin{newitemize}
    \item We develop techniques to modulate visible light that can illuminate an object to cause misclassification on deep learning camera-based vision classifiers, while being completely invisible to humans.  
    Our work contributes to a new class of physical adversarial examples that exploit the differences between human and machine vision. 
    \item We develop a differentiable analytical model of image formation under the radiometric rolling shutter effect and formulate an adversarial objective function that can be solved using standard gradient descent methods.
    \item We instantiate the attack in a physical setting and characterize this new class of attack by studying the effects of camera optics and environmental conditions, such as camera orientation, lighting condition, and exposure. Code is available at~\url{https://github.com/EarlMadSec/invis-perturbations}.
\end{newitemize}

%% file: related.tex
\section{Related Work}\label{sec:relwork}



\paragraph{Digital Adversarial Examples.} This type of attack has been relatively well-studied~\cite{szegedy2014intriguing,goodfellow2014explaining,carlini2017towards,moosavi2015deepfool,papernot2016limitations,biggio2013evasion,kurakin2016adversarial} with several attack techniques proposed. They all involve creating pixel-level changes to the image containing a target object. However, this level of access is not realistic when launching attacks on cyber-physical systems --- an attacker who has the ability to manipulate pixels at a digital level already has privileged access to the system and can directly launch simpler attacks that are more effective. For example, the computer security community has shown how an attacker could directly (de)activate brakes in a car~\cite{carsec}.



\paragraph{Physical Adversarial Examples.}  Physical perturbations are the most realistic way to attack physical systems. Recent work has introduced attacks that require highly visible patterns affixed to the victim object, such  as stickers/patches on traffic signs, patterned eyeglass frames or 3D printed objects~\cite{roadsigns17,athalye2017synthesizing,patch-attack,adv-t-shirt,sharif2016accessorize}. 
We introduce a new kind of physical adversarial example that cameras can see but humans cannot.  
Li et al.~\cite{cam-sticker} recently proposed adversarial camera stickers. These do not require visible stickers on the target object, but they require the attacker to place a sticker on the camera lens. 
By contrast, we target a more common and widely used threat model where the attacker can only modify the appearance of a victim object. 

\paragraph{Rolling Shutter Distortions.} Broadly, rolling shutter can manifest in two kinds of image distortions: (1) motion-based, where the camera or object move during capture, and (2) radiometric, where the lighting varies rapidly during camera exposure. The more common among the two is motion-based, and thus, most prior work has examined techniques to correct motion distortions~\cite{albl2020rolling,geyer_geometric_2005,chia-kai_liang_analysis_2008,bradley_synchronization_2009}. Early works derived geometric models of rolling-shutter cameras and removed image distortions due to global, constant in-plane translation~\cite{geyer_geometric_2005,chia-kai_liang_analysis_2008}, which was later extended to non-rigid motion via dense optical flow~\cite{bradley_synchronization_2009}. Our work focuses on exploiting radiometric distortions caused by high-frequency lights.


\paragraph{Rolling Shutter for Communication.} A line of work has explored visible light communication using the radiometric rolling shutter effect~\cite{disco,rolling-light}. Similar to our work, the goal is to transmit information from a light source to a camera by modulating a high-frequency time-varying light signal such as an LED. We take inspiration from this work and explore how an adversary can manipulate the light source to transmit an adversarial example. However, the key difference is that there is no ``receiver'' in our setting. Rather, the attacker must be able to transmit all information necessary for the attack in a single image without any co-operation from the camera. By contrast, the communication setting can involve taking multiple images over time because the light source and camera co-operate to achieve information transfer. In our case, the light signal must robustly encode information so that the attack effect is achieved in the span of a single image --- a challenge that we address. 

\paragraph{Rolling Shutter for Visual Privacy.} Zhu et al.~\cite{lishield} proposed using radiometric rolling shutter distortions to reduce the signal-to-noise ratio in an image until it becomes unintelligible to humans. This helps to prevent photography in sensitive spaces. Our goal is orthogonal --- we wish to manipulate the rolling shutter effect to cause \emph{targeted} misclassifications in deep learning models. 



%% file: bg.tex
\section{Image Formation under Rolling Shutter} 
\label{sec:bg}

\paragraph{Rolling Shutter Background.} Broadly, cameras are of two types depending on how they capture an image: (1) rolling shutter (RS) and (2) global shutter. A camera consists of an array of light sensors (each sensor corresponds to an image pixel). While an image is being formed, these sensors are exposed to light energy for a period of $\exptime$, known as \emph{exposure time}, and then the data is digitized and read out to memory. In a global shutter, the entire sensor array is exposed at the same time and then the sensors are turned off for the readout operation. By contrast, an RS camera exposes each row of pixels at slightly different periods of time. Thus, while rows are being exposed to light, the data for previously exposed rows are read out. This leads to a higher frame-rate than for high resolution cameras. 


\begin{figure}[t]
    \includegraphics[width=.90\columnwidth]{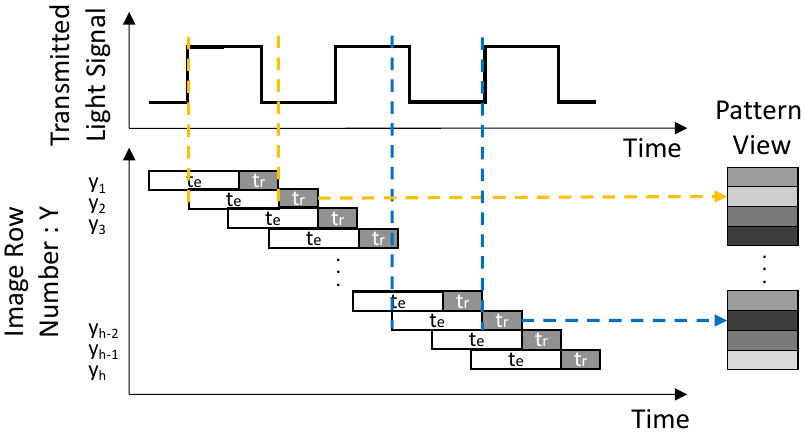}
    \caption{Modulated light induces the radiometric rolling shutter effect. Here
      $\readouttime$ denotes the time it takes to read a row of sensors, and
      $\exptime$ denotes the exposure of the camera.}
      
    \label{fig:rs}
\end{figure}

We visualize the rolling shutter effect in the presence of lighting changes
in~\figref{fig:rs}. For an RS camera, the time it takes to
read a row is called readout time ($\readouttime$).\footnote{This is also approximately
  the time difference between when two consecutive rows are exposed.}  Each
row is exposed and read out at a slightly later time than the previous row. Let
$t_0$ be the time when the first row is exposed, then the $y\thh$ row is exposed
at time $t_0 + (y-1)\readouttime$, and read at
$t_0 + (y-1)\readouttime + \exptime$.

As different rows are exposed at different points in time, any lighting or
spatial changes in the scene that occurs while the image is being taken 
can lead to undesirable artifacts in the captured image, including distortion or
horizontal stripes on the image, known as \emph{rolling shutter
  effect}~\cite{liang2008analysis}. 
In this work, we exploit such artifacts by
modulating a light source. 
We contribute a technique to determine the precise
modulation required to trick state-of-the-art deep learning models for visual
classification.

\paragraph{Image Formation.} 
We represent the time-modulated attacker signal as
$\signal(t)$. We assume that the scene contains ambient light in addition to the
attacker-controlled light source (e.g., a set of Smart LED lights). Let
$\ltex(x,y)$ represent the texture of the scene, which we approximate as the
value of the $(x, y)$ pixel. 
As the attacker signal is a function of time, the illumination at pixel $(x, y)$ on the
scene will vary over time, $(\alpha + \beta\:\signal(t))$. 
Here $\alpha$ and $\beta$ represent the intensity of the ambient light and the
maximum intensity of the attacker controlled light, respectively.  
We note that the
attacker can use an RGB LED, and thus, the attacker's signal contains three
components: Red, Green and Blue.
   



In rolling shutter camera, pixels on the same row
are exposed at the same time, and neighboring rows are exposed at slightly
different times. Let each row be exposed for $\exptime$ seconds, and the $y\thh$
row starts exposing at time $t_y$.  Therefore, the intensity of a pixel
$(x, y)$ in row $y$, will be: $i(x,y) = \gain \int_{t_y}^{t_y+\exptime} \ltex(x,y)\:\left(\alpha + \beta\:\signal(t)\right) \: dt$.
Here, $\gain$ denotes the sensor gain of the camera sensor that converts the light
radiance falling on a pixel sensor into a pixel intensity. Thus, we have:
\bea %
    i(x,y)
    & = & \gain\:\ltex(x,y)\left(\alpha\exptime +  \beta\int_{t_y}^{t_y+\exptime}  \signal(t)\: dt\right) \\
    & = & \gain\:\ltex(x,y)\:\exptime\:\alpha +  \gain\:\ltex(x,y)\:\exptime\:\beta\: g(y)\\
    & = & \ambimg + \sigimg\cdot g(y)
\eea %
\noindent Here, the signal image $g(y)$ denotes the average effect of signal $\signal(t)$ on row
$y$, $g(y) = {1\over\exptime}\int_{t_y}^{t_y+\exptime} \signal(t)\: dt$.  Let
$\ambimg$ be the image captured under only ambient light, such that
$\ambimg = \gain\:\ltex(x,y)\exptime\alpha$, and $\sigimg$ is the image captured under
only the full illumination of the attacker controlled light (with no ambient light).  

The
time-varying signal $\signal(t)$ we generate is periodic, with period $\period$;
during the image capture the signal could have an offset of $\delta$ with
respect to the camera. Therefore, final equation of pixel intensity would be,
\bne
\finalimg = \ambimg + \sigimg\cdot g(y + \delta)
\label{eq:final}
\ene
In the next section, we discuss how we make our attack robust to environmental conditions, including any offset $\delta$.

%% file: attack-theory.tex
\section{Crafting Invisible Perturbations} 
\label{sec:theory}

\begin{figure}[t]
    \centering
    \includegraphics[width=1.0\linewidth]{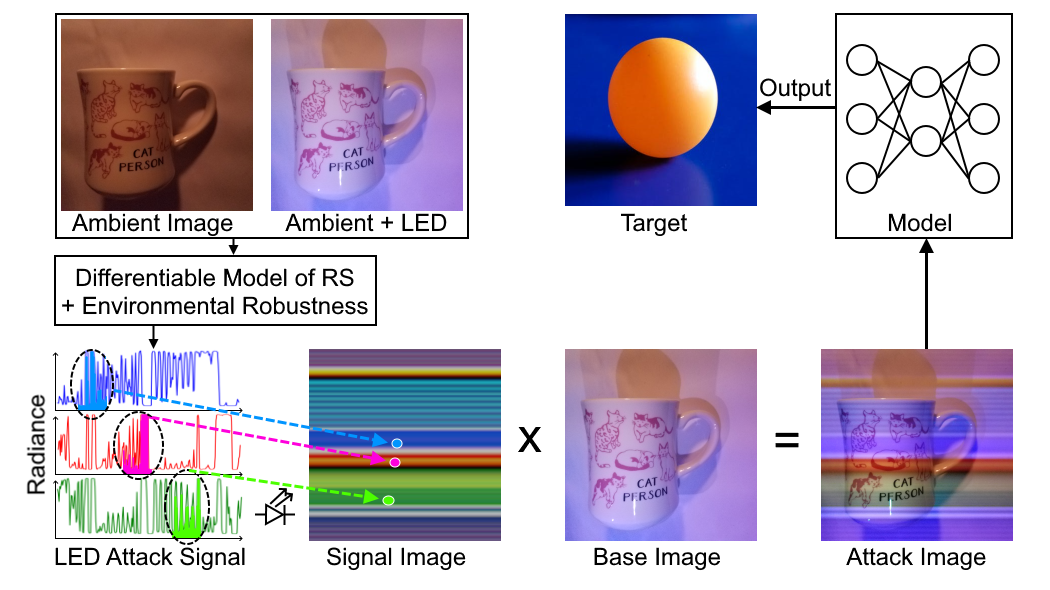}
    \caption{The attacker creates a time-modulated high frequency light signal that induces radiometric striping distortions in rolling shutter cameras.  The striping pattern is designed to cause misclassifications.}
    \label{fig:pipeline}
\end{figure}

Our high-level goal is to generate a light signal by modulating a light source
such that it induces striping patterns when a rolling shutter camera senses the
scene. These patterns should be adversarial to a machine learning model but
should not be visible to humans. The attacker light source flickers at a
frequency that humans cannot perceive, and thus, the scene simply appears to be
illuminated. \figref{fig:pipeline} outlines the attack pipeline. 
To achieve this goal, we first present the challenges in crafting
such light modulation, followed by our algorithm for overcoming these issues.

\subsection{Physical World Challenges}
\label{subsec:phy-challenges}

One of the key challenges in creating physical adversarial examples is to
create a simulation framework that can accurately estimate the final image taken
by the camera. Without such a framework it will be very slow to compute an
attack by repeating physical experiments.
In addition, physical world perturbations must survive
varying environmental conditions, such as viewpoint and lighting changes. Prior
work has proposed methods that can create
adversarial examples robust to these environmental factors. However, in our
setting, we encounter a unique set of additional challenges concerning light
generation, reception, and camera optics. 

\paragraph{Desynchronization between camera and light source.}
The location of the striping patterns appearing on the image depends on the
synchronization between the camera and the light source. Failing to do so, will cyclically permute the striping pattern on the image, resulting in a different final image.
However, the attacker has no control
over the camera and when the image is taken.  Therefore, 
we optimize our signal to remain adversarial even when the light source is out
of sync with the camera at image capture time.


\paragraph{Camera exposure.}
The exposure of the camera will significantly change how a particular attacker
signal is interpreted.  A long exposure will apply a ``smoothing effect''
on the signal as two consecutive rows will receive much of the same light.  This
will reduce the attacker's ability to cause
misclassifications. 
A shorter exposure would create more pronounced bands on the image, making it
easier to induce misclassification.  We show that our adversarial signal can be
effective for a wide range of exposure values. 

\paragraph{Color of light production and reception.}  Prior work has examined fabrication
error in the case of printer
colors~\cite{roadsigns17,sharif2016accessorize}. Our attack occurs through an
LED and this requires different techniques to account for fabrication errors:
(1) 
Red, Green, Blue LEDs produce light of different intensities; (2) Cameras run
proprietary color correction; (3) Transmitted light can bleed into all three
color channels (e.g., if only Red light is transmitted, on the sensor side, it
will still affect the Green and Blue channels).  We learn approximate functions
to translate a signal onto an image so that we can create a simulation framework
for quickly finding adversarial examples.


\subsection{Optimization Formulation}
\label{subsec:opt}
Our goal is to compute a light signal $\signal(t)$ such that, when an image is
taken under the influence of this light signal, the loss is minimized between
the model output and the desired target class.
However, unlike prior formulations, we do not need an $\ell_P$ constraint on
perturbation magnitude because our perturbations (via high-frequency light
modulation) are invisible to human eyes by
design. 
Instead, our formulation is constrained by the capabilities of the LEDs, the
Arduino chip we use to modulate them (see~\secref{sec:realize}), and the camera
parameters. 
A novel aspect in our formulation is the differentiable representation of the
rolling shutter camera and color correction applied by the camera. Such
representation allows us to compute the adversarial example end-to-end using
common gradient-based methods, such as PGD~\cite{madry2017towards} and
FGSM~\cite{goodfellow2014explaining}. Our model allows us to manipulate camera parameters such as exposure time, image size and row readout rate.

Following~\eqref{eq:final}, we get the final image
$\finalimg$ as a sum of the image in ambient light($\ambimg$) and in only the attacker's light source($\sigimg$). Based on the image formation model discussed above, we have the following objective
function:
%
%
\begin{equation} \label{objfunc}
\begin{aligned}
  &\min_{f(t)} \:\mathop{\Exp}_{\colorcorr,\viewtransform,\delta} J\left(\model(\colorcorr(\finalimg)), \targetclass\right) \\
  &\finalimg = \viewtransform\left(\ambimg\right) + \viewtransform\left(\sigimg\right)\cdot g(y+\delta)\\
  & g(y) = {1\over\exptime}\int_{t_y}^{t_y+\exptime} f(t)\: dt
\end{aligned}
\end{equation}
where $J(.,k)$ is the classification loss for the target class $\targetclass$,
$\model$ is the classifier model, $\colorcorr$ is a function to account for
color reproduction error, 
$\viewtransform$
models viewpoint and lighting changes, $\delta$ 
denotes possible signal offsets. 
The image under only ambient light is $\ambimg$ and under only fully illuminated
attacker-controlled light is $\sigimg$. 

As we assume the attacker does not have control over the ambient light, we
cannot take $\sigimg$ (image without the effect of the ambient light). We
instead take an image where both ambient and the attacker controlled LEDs are
fully illuminated, which we call $\fullimg = \ambimg + \sigimg$, and extrapolate
$\sigimg$ as $\fullimg - \ambimg$.

%

The process of solving the above optimization problem is shown
in~\algref{algo:gen-signal}. We use the cross-entropy as our loss function $J$
and used ADAM~\cite{kingma2014adam} as the optimizer.
Next, we discuss how our algorithm handles the unique challenges
(\secref{subsec:phy-challenges}) to generate robust adversarial signals. 

\newcommand{\varw}{v}
\newcommand{\sigsize}{l}
\begin{algorithm}[t]
  \gamesfontsize
  \textbf{Input:} Image with only ambient light $\ambimg$, image with ambient and attacker controlled lights $\fullimg$,  target class $\targetclass$, and exposure value $\exptime$\\
  \textbf{Output:} Digitized adversarial light signal $\digsignal$, which is an vector of size $\sigsize$. \\
  \textbf{Notations:} $c$: number of color channels; $\shift(.,\delta)$: cyclic
  permutation of an vector shifted by $\delta$ places; $\gamma$: parameter for
  gamma correction; $N$: threshold for maximum number of iteration; $s$ is the
  shutter function which depends on the $\exptime$ and image size $h\times w$
\begin{algorithmic}
\Procedure{OPTIMIZE}{$\ambimg, \fullimg, k, s$}
\State $n\gets 1$
\State $\varw_{0} \getsr \Z^{c\times \sigsize}$  \Comment{Randomly sample an vector of size {$c\times \sigsize$}}
\While{not converge and $n\le N$}
    \State $\colorcorr \sim P, \viewtransform \sim X, \delta \sim \{0,1, \ldots, \sigsize\}$
    \State $\digsignal_n \gets \frac{1}{2} (\tanh(\varw_{n-1})+1)$
    \State $o_n \leftarrow$ $\shift(\digsignal_n,\delta)$ \Comment{cyclic permutation}
    \State $g_n \leftarrow o_n * s$ \Comment{convolution with the shutter function}
    \State $\ambimgn \gets \viewtransform(\ambimg); \;\; \fullimgn \gets \viewtransform(\fullimg)$
    \State $\sigimgn \gets \left(\ambimgn^{\gamma} + g_n \times (\fullimgn^{\gamma} - \ambimgn^{\gamma})\right)^{1 \over \gamma}$
    \State $L \gets J\left(\model\left(\colorcorr(\sigimgn)\right), \targetclass\right)$ \Comment{loss for target class k}
    \State $\Delta \varw \leftarrow \nabla_{\varw_{n-1}} L$
    \State $\varw_n \leftarrow \varw_{n-1} + \Delta \varw$
    \State $n \leftarrow n + 1$
\EndWhile
\State $\digsignal \gets \frac{1}{2} (\tanh(\varw_n)+1)$
\State \textbf{return} $\digsignal$
\EndProcedure
\end{algorithmic}
 \caption{Adversarial Light Signal Generation}
 \label{algo:gen-signal}
\end{algorithm}

\paragraph{Structure of $\signal(t)$.} One of the challenges in solving the
above optimization problem is determining how to represent the time-vary
attacker signal $\signal(t)$ in a suitable format. We choose to represent it as
an vector of intensity values, which we denote as $\digsignal$. Each index in
$\digsignal$ represents a time interval of $\readouttime$ (i.e., the readout
time of the camera). This is because the attacker will not gain any additional
control over the rolling shutter effect by changing the light intensity within a
single $\readouttime$ period: Within a single $\readouttime$, the same set of
rows are exposed to light and any intensity changes will be
averaged. Furthermore, we bound the values of $\digsignal$ to be in $[0,1]$,
such that $0$ denotes zero intensity and $1$ denotes full intensity. The signal values inside are scaled accordingly. 
To ensure our signal is within the bounds, we use a change-of-variables.
We 
define $\digsignal = \frac{1}{2}(\tanh(\varw) + 1)$. Thus, $\varw$ can take any
unbounded value during our optimization. Finally, the attacker must determine
what is an appropriate length of $\digsignal$ because the optimizer needs a
tensor of finite size. We design $\digsignal$ to be periodic with period equal
to image capture time: $\readouttime \cdot h + \exptime$ where $h$ is the height
of the image in pixels.  As each index in $\digsignal$ represents $\readouttime$
units of time, the length of the vector for $\digsignal$ would be
$\sigsize = h + \Bigl \lceil{{\exptime}\over \readouttime}\Bigl \rceil$.  



\paragraph{Viewpoint and Lighting Changes.} We build on prior work in obtaining robustness to viewpoint (object pose) and lighting variability. Specifically, we use the expectation-over-transformation approach (EoT) that samples differentiable image transformations from a distribution (E.g., rotations, translations, brightness)~\cite{athalye2017synthesizing}. We model this using distribution $X$ which consists of transformations for flipping the image horizontally and vertically, magnifying the image to account for small distance variations, and planar rotations of the image. During each iteration of the optimization process, we sample a transformation $\viewtransform$ from $X$ and apply it to the pair of object images $\ambimg$ and $\fullimg$.  We apply multiplicative noise to the ambient light image $\ambimg$ to model small variations in the ambient light. However, to account for a wider variation in the ambient light, we adjust our signal during attack execution. This is one of the key benefits of this attack to be agile to environment changes. We generate a set of adversarial light signals, each designed to operate robustly at specific intervals of ambient light values. During the attack, we switch our light signal to the one that corresponds to the current ambient light setting.\footnote{The attacker could measure the approximate ambient light using a light meter attached to the attacker controlled light, e.g.~\url{https://www.lighting.philips.com/main/systems/themes/dynamic-lighting}.} Using this approach, we avoid optimizing over large ranges of ambient light conditions and hence, improve the effectiveness of our attack.

\paragraph{Signal Offset.} Because our signal can have a phase difference with
the camera, we account for this during optimization. The offset is an integer
value $\delta \in \{0, 1, \ldots, \sigsize\}$. Each offset value can be
represented by a specific cyclic permutation of the $\digsignal$ vector. A offset
value of $\delta$ corresponds to performing a $\delta$-step cyclic rotation on
the signal vector. To gain robustness against arbitrary offsets, we model
the cyclic rotation as a matrix multiplication operation. This enables us to use
EoT by sampling random offsets during optimization.



\paragraph{Color Production and Reception Errors.}
Imperfections in light generation and image formation by the camera can lead to 
errors. Furthermore, the camera can run proprietary correction steps such as gamma correction to improve image quality. We account for the gamma correction by using the sRGB (Standard RGB) standard value, $\gamma = 2.2$~\cite{srgbstandard}. However, it is infeasible to model all possible sources of imperfection. Instead, we model the fabrication error as a distribution of transformations in a coarse-grained manner and perform EoT to overcome the color discrepancy. The error transformations are a set of experimentally-determined affine ($Ax+B$) or polynomial ($a_0 x^n+a_1x^{n-1}+...+a_n$) transformations applied per color channel (term $\colorcorr$ in~\eqref{objfunc}). Please see the supplementary material for exact parameter ranges for the distribution $P$ from which we sample $\colorcorr$ values.

\paragraph{Handling Different Exposures.}
Eq.~\ref{objfunc} models the effect of the attacker signal on the image as a convolution between $\signal(t)$ and a shutter function. Shorter exposure leads to smaller convolution sizes, and longer exposure leads to larger convolution size. 
Instead of optimizing for different exposure values, 
we take advantage of a feature of this new style of physical attack --- its dynamism. Specifically, the attacker can optimize different signals $f(t)$ for different discrete exposure values and then, at attack execution, switch to the signal that is most appropriate to the camera being attacked and ambient light. As most cameras have standard exposure rates, the attacker can \emph{apriori} create different signals. We note that dynamism is a feature of our work and is not possible with current physical attacks~\cite{roadsigns17,athalye2017synthesizing,cam-sticker,adv-t-shirt,sharif2016accessorize,patch-attack}. 

%% file: attack-physical.tex
\section{Producing Attack Signal using LED lights}
\label{sec:realize}
We used a simulation framework to generate adversarial light signals for a given
scene and camera parameters. To validate that these signals are effective in the real world, we implement the attack using programmable LEDs. The primary challenge we address here is modulating an LED according to the optimized signal $\digsignal$, a vector of reals in $[0, 1]$.




We use an Arduino Atmel Cortex M-3 chip (clock rate $84$~MHz) to drive a pair of RGB LEDs.\footnote{MTG7-001I-XML00-RGBW-BCB1 from Marktech Optoelectronics.} We used a Samsung Galaxy~S7 for taking
images, whose read out time ($\readouttime$) is around $10~\mus$.  
The camera takes
images at resolution $3024\times 3024$, which is $12$x larger than the input size that our algorithm requires $252\times252$. Our optimization process resizes images to $224\times224$ before passing to ResNet-101 
classifier. 
Thus, when a full-resolution image is resized to the dimensions of the model, 12 rows of data get resized to 1 row. We account for this by defining an effective readout time of $120~\mus$. That is, the LED signal is held for $120~\mus$ before moving to the next value in $\digsignal$. Recall that we do not need to change the signal intensity within the readout time because any changes during that time will be averaged by the sensor array.


We drive the LEDs using pulse width modulation to produce the intensities specified in the digital-version of attack signal $\digsignal$. Driving three channels simultaneously with one driver requires pre-computing a schedule for the PWM widths. This process requires fine-grained delays, so we use the \textsf{delayMicroseconds} function in the Arduino library that provides accurate delays greater than $4~\mus$. The attack might require delays smaller than this value, but it occurs rarely and does not have an effect on the fabricated signal (\secref{sec:expts}).\footnote{There can be a small difference between the
  period for duty cycle and the camera readout time ($\readouttime$). But as our
  exposure rate $\exptime >= 0.5~ms$ is significantly larger
  than row readout time $\readouttime=10~\mus$, this difference has only little
  affect on our attack.} 

%% file: exp.tex
\section{Experiments}
\label{sec:expts}

We experimentally characterize the simulation and physical-world performance of adversarial rolling shutter attacks. For all experiments, we use a ResNet-101 classifier trained on ImageNet~\cite{deng2009imagenet}. The experiments show that:
(1) We can induce consistent and targeted misclassification by modulating lights that is robust to camera orientation. (2) Our simulation framework closely follows physical experiments, therefore the signals we generate in our simulation also translate to robust attack in physical settings; (3) The effectiveness of the attack signal depends on the camera exposure value and ambient light --- longer exposure or bright ambient light can reduce attack efficacy.  

For evaluating each attack, we take a random sample of images with different signal phase
shift values ($\delta$) and viewpoint transformations ($\viewtransform$). We define attack accuracy
as the fraction of these images classified as the target. We also record the
average classifier's confidence for all the images when it is classified as the
target.  



\begin{table}[t]
\begin{center}\tabfontsize
\begin{tabular}{p{0.4in}p{0.6in}cc}
  \toprule
Source (confid.) & Affinity targets & \multicolumn{1}{p{0.3in}}{Attack success} & \multicolumn{1}{p{0.7in}}{Target confidence (StdDev)} \\
\midrule
\multirow{3}{0.5in}{Coffee mug (83\%)} 
  & Perfume & 99\% & 82\% (13\%)\\
  & Candle & 98\% & 85\% (18\%)\\
  & Ping-pong ball & 79\% & 68\% (27\%)\\\midrule
\multirow{3}{0.5in}{Street sign (87\%)} & Monitor       & 99\% & 94\% (12\%)\\
  & Park bench & 99\% & 90\% (13\%)\\
  & Lipstick & 84\% & 78\% (20\%)\\\midrule
\multirow{3}{0.5in}{Soccer ball (97\%)}
  & Pinwheel      & 96\% & 87\% (15\%)\\
  & Goblet      & 78\% & 55\% (17\%)\\
  & Helmet      & 66\% & 59\% (22\%)\\\midrule
\multirow{3}{0.5in}{Rifle (96\%)} & Bow & 76\% & 64\% (24\%)\\
  & Tripod              & 65\% & 65\% (22\%)\\
  & Binoculars              & 35\% & 40\% (18\%)\\\midrule
\multirow{3}{0.5in}{Teddy bear (93\%)} & Tennis ball    & 92\% & 88\% (19\%)\\
  & Acorn    & 75\% & 72\% (25\%)\\
  & Eraser    & 47\% & 39\% (16\%)\\
\bottomrule
\end{tabular}
\end{center}
\caption{Performance of affinity targeting using our adversarial light signals
  on five classes from ImageNet. For each source
  class we note the top 3 affinity targets, their attack success rate, and
  average classifier confidence of the target class. (Average is taken
    over all offsets values for 200 randomly sampled transformations.)}
\label{tab:simulation-results}
\end{table}

\subsection{Simulation Results}
\label{subsec:sim-results}

For understanding the feasibility of our attack in simulation we selected five
victim objects. As our signal crafting process requires two images --- object
under ambient light and object with LEDs at full capacity --- we approximate the
image pair by adjusting the brightness of the base image present in ImageNet
dataset.  For $\ambimg$, we ensure the average pixel intensity is $85$ (out of
$255$) and for $\fullimg$ it is $160$. Both values are chosen to mimic what we
get in our physical experiments. 
Then, we optimize for
various viewpoints using the EoT approach.

As light-based attacks have a constrained effect on the resulting image (i.e.,
translucent striping patterns where each stripe has a single color) compared to
current physical attacks, we found that it is not possible to randomly select
target classes for the attack. Rather, we find that certain target classes are
easier to attack than others. We call this \emph{affinity
  targeting}. Concretely, for each source class, we compute a subset of affinity
targets by using an untargeted attack for a small number of iterations (e.g.,
$1000$), and then pick the top~10 semantically-far target classes --- e.g., for ``coffee mug,'' we ignore targets like ``cup'' --- based on the classifier's
confidence.  
Then, we use targeted attack using the affinity targets. The results are shown
in~\tabref{tab:simulation-results}. For brevity, we show three affinity
targets for each source class. (Please see the supplementary material for full results.)

\subsection{Physical Results}
\label{subsec:phy-results}
\begin{figure}[t]
 \centering
 \includegraphics[height=2.7cm]{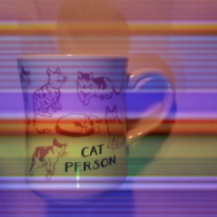}
 \includegraphics[height=2.7cm]{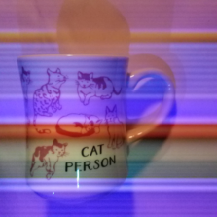}
 \caption[realizable]{The simulation framework closely replicates the radiometric rolling shutter effect.  
 The left image shows the simulation result and the
   right one is obtained in the physical experiments. Both of them are classified as
   ``ping-pong ball.'' 
 }
 \label{fig:sim-phy}
\end{figure}

We characterize the attack algorithm's performance across various camera configurations and environmental conditions. 
We find that the physical world results generally follow the trend of
simulation results, implying that computing a successful simulation result will
likely lead to a good physical success rate. \figref{fig:sim-phy}
confirms that the simulated image is visually similar to the physical one. 
To ensure the baseline imaging condition is valid, for all physical
testing conditions, we capture images of the victim object under the same
exposure, and similar ambient light and viewpoints. All of the baseline images 
 are correctly classified as the object (e.g., coffee mug) with an average
confidence of $68\%$. 


\begin{figure*}[!t]
\centering
\includegraphics[height=0.25cm]{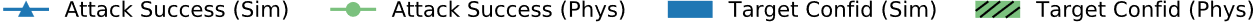}\vspace{0.001cm}
\subfloat[
Exposure]
{
  \includegraphics[height=3.6cm]{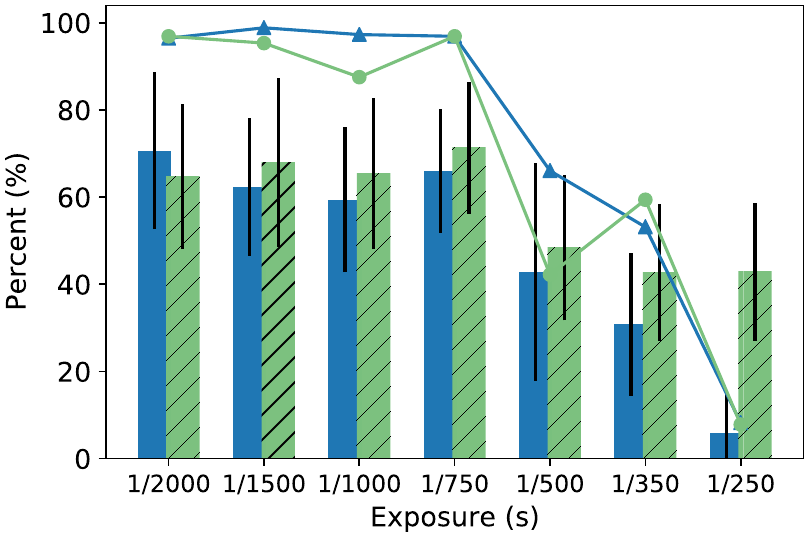}%
\label{fig:conv-exposure}}
\hfill
\subfloat[
Ambient light intensity]
{\includegraphics[height=3.6cm]{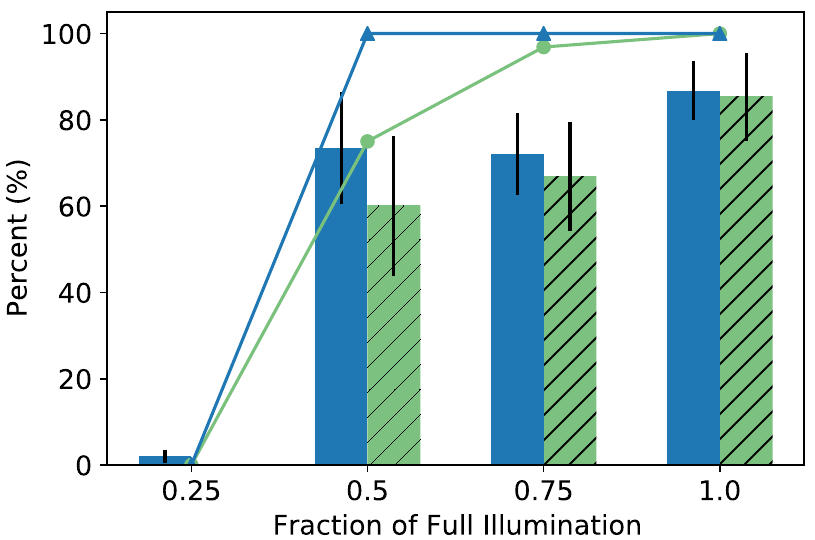}%
\label{fig:ambient-light}}
\hfill
\subfloat[
Field of View (FoV)]{\includegraphics[height=3.6cm]{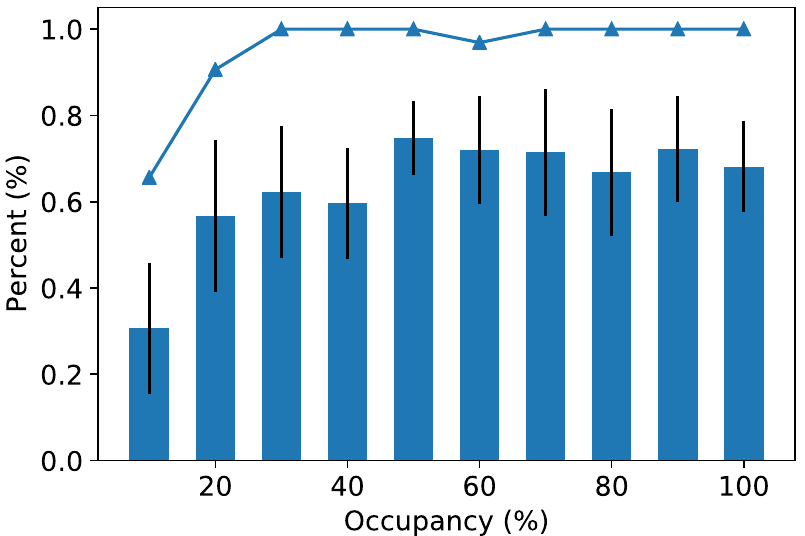}%
\label{fig:fov}}
\caption{
Evaluating the attack success rate in simulation (\texttt{Sim}) and physical (\texttt{Phys}) for different settings (such as, ambient lighting and field-of-view) and camera parameters (such as, exposure).
}
\label{fig:phy_exp}
\end{figure*}

\paragraph{Effect of Exposure.}
We first explore the range of camera exposure values in which our attack would be effective. \figref{fig:conv-exposure} shows the effect of various common exposure settings on the attack's efficacy. 
We observe that the attack performs relatively well --- approximately $94\%$
targeted attack success rate with $67\%$ confidence --- at exposures $1/750$s
and shorter. However, as exposures get longer the efficacy of the attack
degrades and it stops working at exposures longer than $1/250$\textit{s}. This confirms
our hypothesis that longer exposures begin to approximate the global shutter
effect. 
Based on the exposure results, we select a setting
of $1/2000$\textit{s} for the following experiments. 



\paragraph{Ambient Lighting.} Attack performance depends on the lighting
condition. We have experimentally
observed that EoT under widely-varying lighting conditions does not converge for
our attack. We emulate different ambient light conditions by controlling the LED
output intensity as a fraction of total ambient lighting. We compute different signals for different ambient light condition and show their attack efficacy at an exposure of $1/2000$\textit{s} in~\figref{fig:ambient-light}. 
As expected the attack performs better as relative strength of LEDs compared to the ambient light is higher.

\paragraph{Various Viewpoints.} We apply EoT to make our signal robust to
viewpoint variations.  In~\figref{fig:viewpoint} (row 1-2), we show the resulting images
with our light signal for different camera orientations and distances for two
different exposure values. All images are classified as ``perfume''.  Physical
targeted attack success rate is $84\%$ with average confidence of $69\%$ at an
exposure of $1/2000$\textit{s}, and a success rate of $72\%$ with average confidence of
$70\%$ at an exposure of $1/750$\textit{s}. The averages are computed across $167$ and
$194$ images at varying camera
orientations. 
In ~\figref{fig:viewpoint} (row 3), we demonstrate the attack against a different object.

\paragraph{Field of View (FoV).}
We optimize attack signals for different FoV occupancy values --- the fraction of foreground object pixels to the whole image --- and observe, in simulation, that the attack is stable until FoV occupancy $\le10\%$ (Fig.~\ref{fig:fov}). In the baseline case, the object is correctly classified at all FoV occupancy values, but the confidence reduces to 51\% when FoV occupancy is $\le10\%$.

\newcommand{\imgsz}{2cm}
\begin{figure}[t]
\setlength{\belowcaptionskip}{-10pt}
  \centering
  \includegraphics[height=\imgsz]{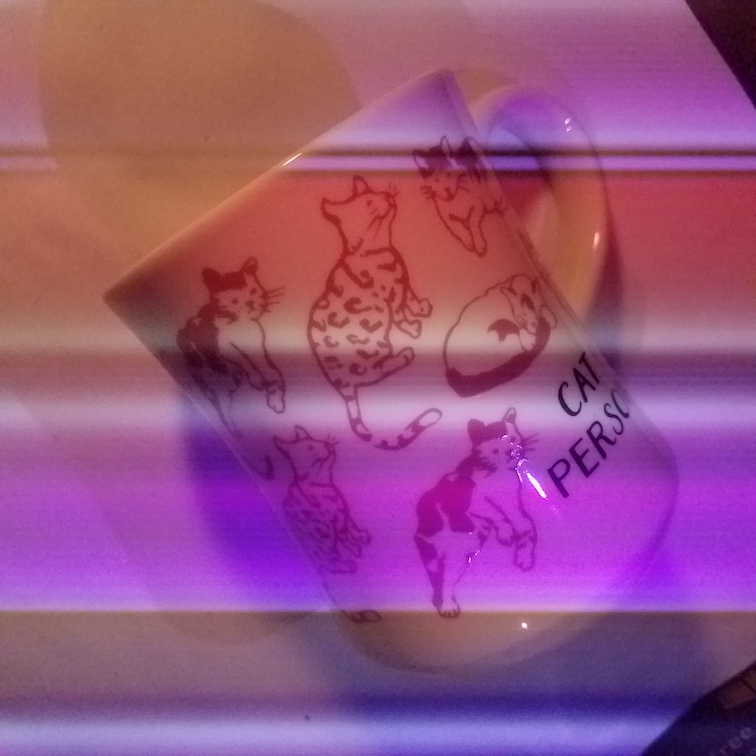}
  \includegraphics[height=\imgsz]{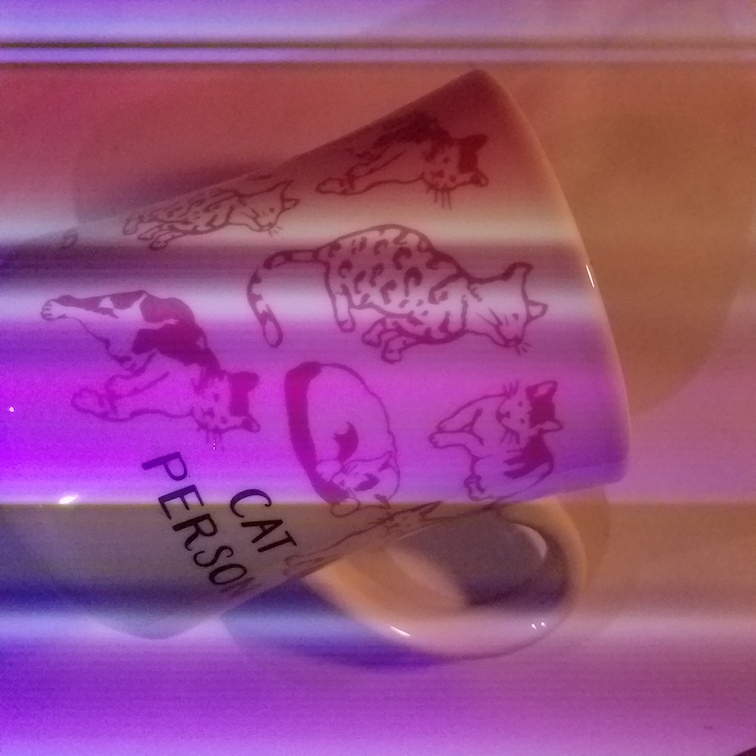}
  \includegraphics[height=\imgsz]{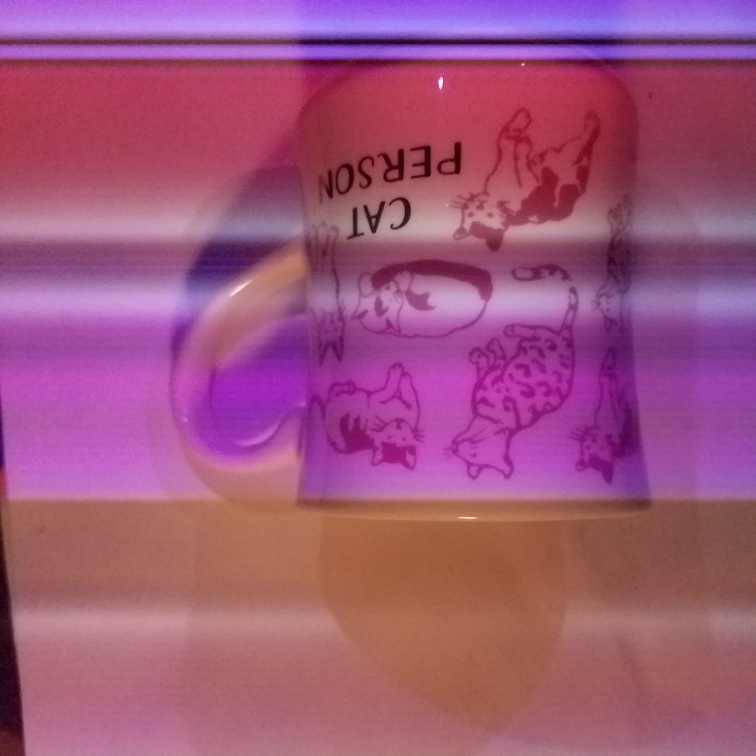}
  \includegraphics[height=\imgsz]{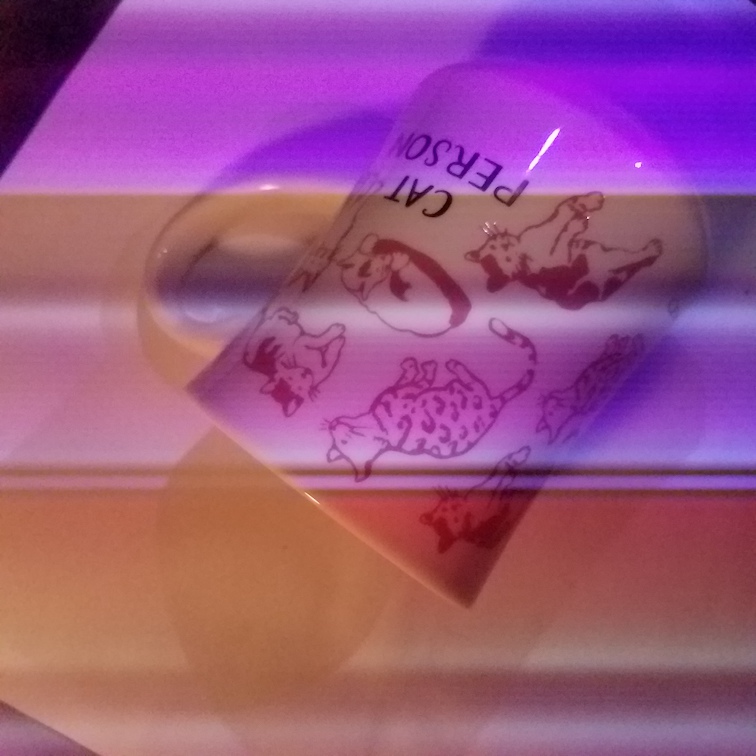} \\
  \includegraphics[height=\imgsz]{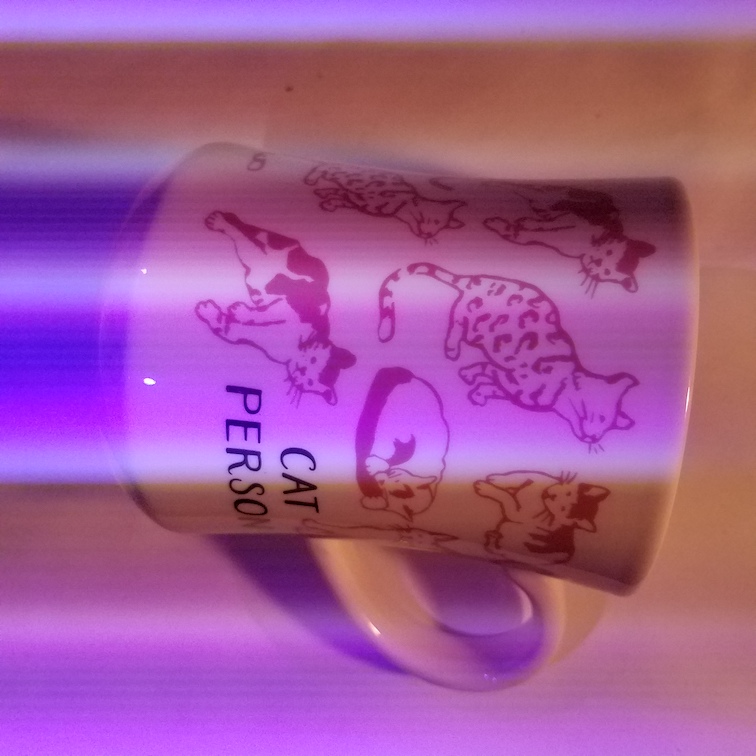}
  \includegraphics[height=\imgsz]{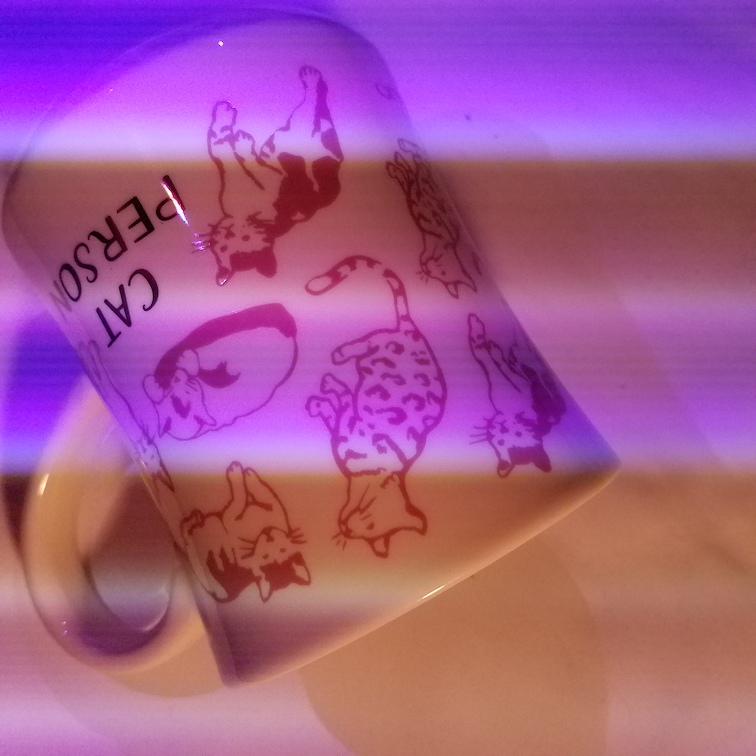}
  \includegraphics[height=\imgsz]{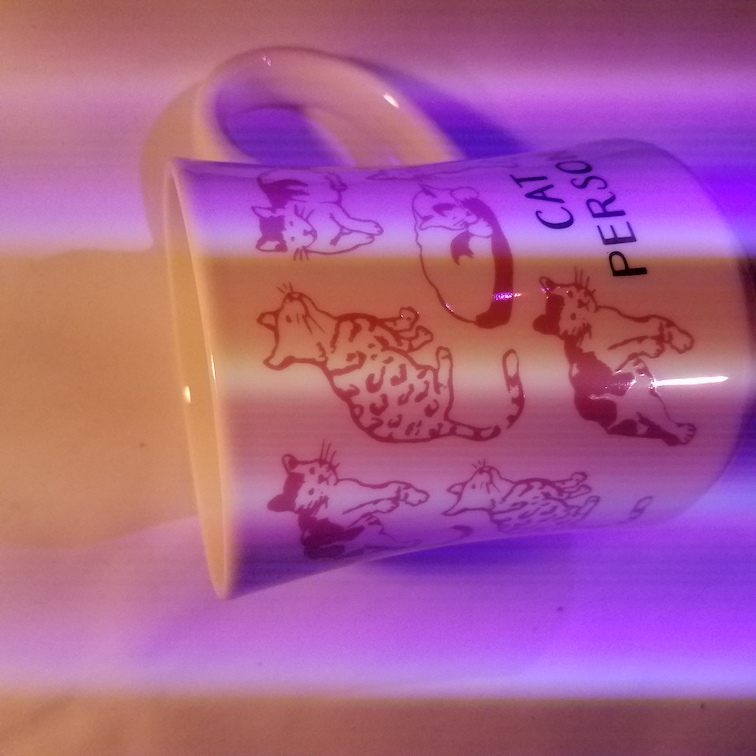}
  \includegraphics[height=\imgsz]{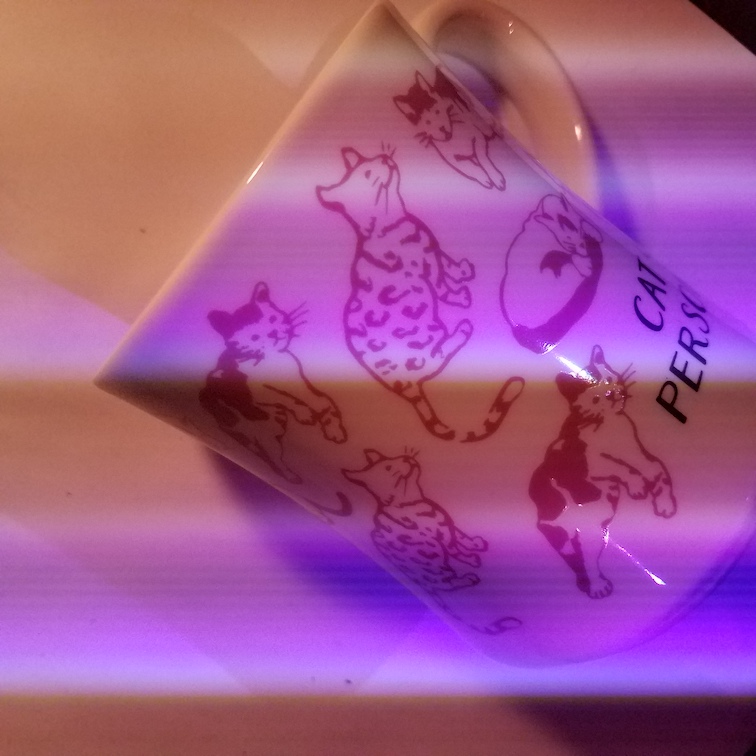}
  
  \includegraphics[height=\imgsz]{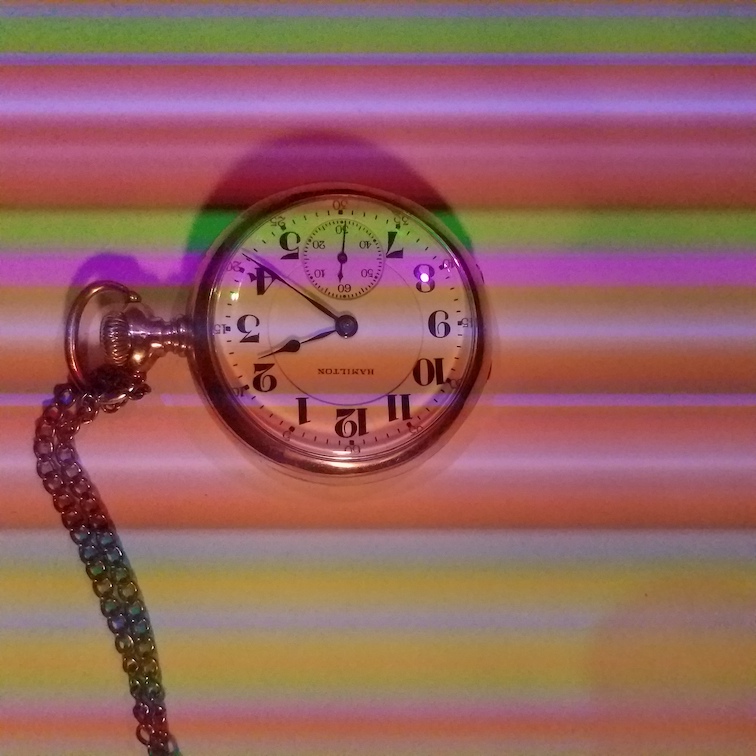}
  \includegraphics[height=\imgsz]{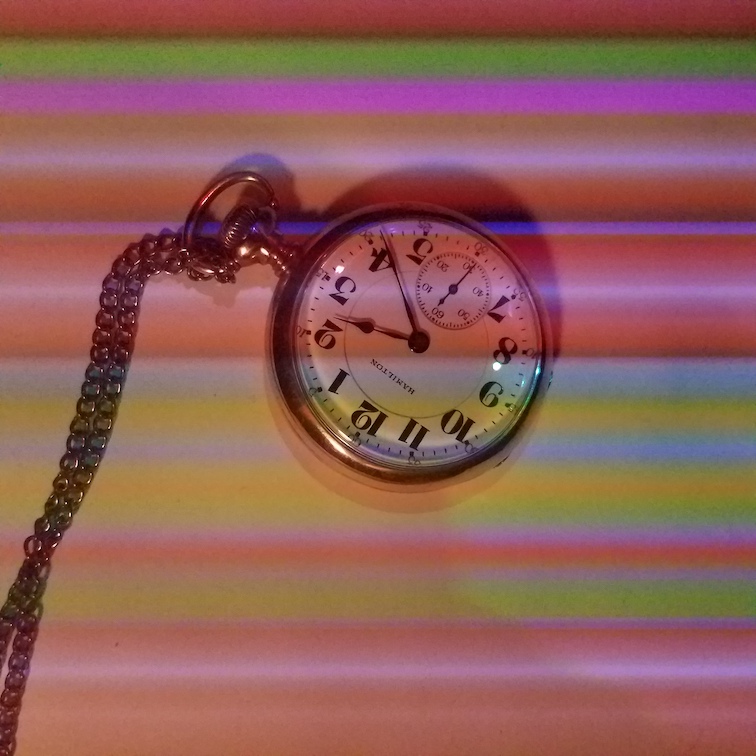}
  \includegraphics[height=\imgsz]{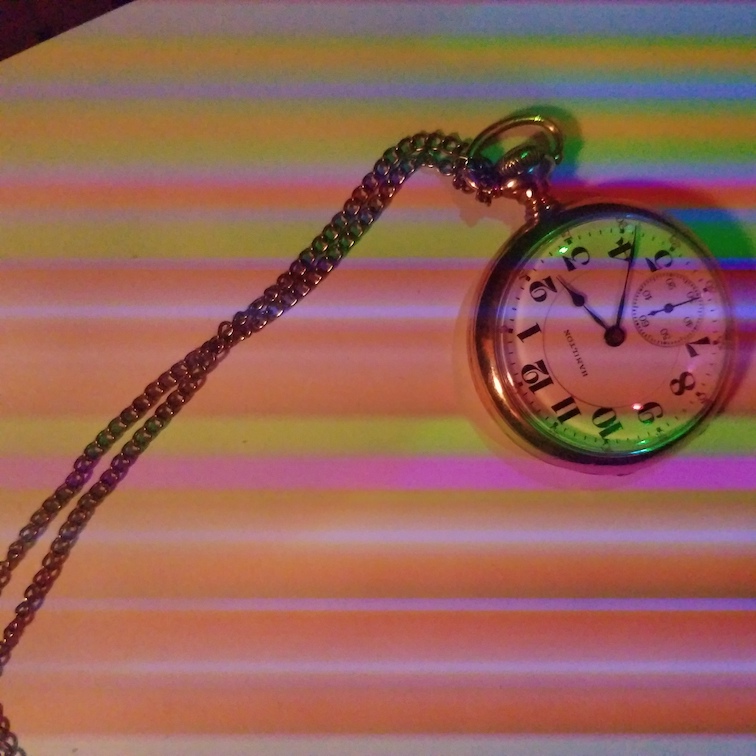}
  \includegraphics[height=\imgsz]{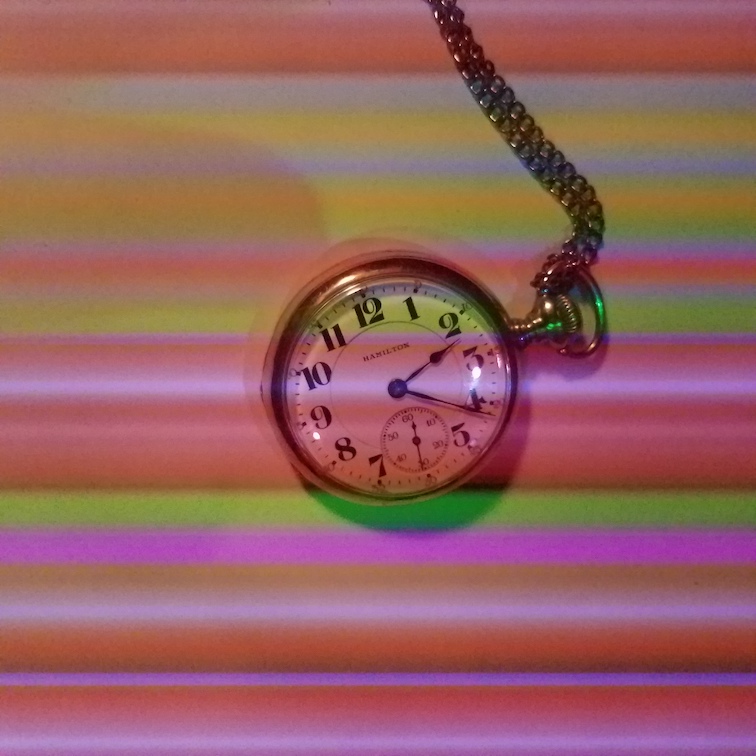}
  
  \caption[viewpoint]{A sample of images taken at different camera
    orientations and two exposure values, $1/2000$\textit{s} (first row) and $1/750$\textit{s}
    (second row). Two different signals are used which are optimized for
    respective exposure
    values. 
    The images are classified as ``perfume'' at an accuracy of $86\%$ (for
    exposure of $1/2000$\textit{s}) and $72\%$ (for exposure of $1/750$\textit{s}) with an average
    confidence of $69\%$.  
    Third row - The images are classified as ``whistle'' at a targeted-attack success rate of $79\%$ with an average confidence of $66\%$.}

  \label{fig:viewpoint}
\end{figure}

%% file: discussion.tex
\section{Discussion and Conclusion}\label{sec:discussion}


\paragraph{High frequency ambient sources.}
For low exposure settings, ambient light sources powered by alternating current (AC) can induce their own flicker patterns~\cite{flicker}. This results in a sinusoidal flicker with a time period that depends on the frequency of the electric grid, which is generally 50Hz or 60Hz. We can address this in our imaging model by adding a signal image component to the ambient image, and use EoT to generate an attack that is invariant to this interference.

\paragraph{Deployment.}
We envision the attack being deployed in low-light or controlled indoor lighting situations. For example, an attacker might compromise a LED bulb in a home to evade smart cameras or face recognition on a smart doorbell or laptop. Here, the attacker can acquire prior knowledge of the sensor parameters (e.g., they can purchase a similar device or lookup specs on the Internet). Given this knowledge, the attacker can pre-optimize a set of signals for commonly occurring imaging conditions for their use-case, measure the situation at deployment time and emit the appropriate signal.

%% file: conclusion.tex
\paragraph{Summary.}
We create a novel way to generate physical adversarial examples that do not
change the object, but manipulate the light that illuminates it. By modulating light at a frequency higher than human perceptibility, we show how to create an
invisible perturbation that rolling shutter cameras will sense and the resulting
image will be misclassified to the attacker-desired class.
The attack is dynamic because an attacker can change the target class or gain robustness against specific ambient lighting or camera exposures by changing the modulation pattern on-the-fly. Our work contributes to the growing understanding
of physical adversarial examples that exploit the differences in machine and human vision. 

\vspace{0.3cm}
{\small
\noindent\textbf{Acknowledgements.} This work was supported in part by the University of
Wisconsin-Madison Office of the Vice Chancellor for Research and Graduate
Education with funding from the Wisconsin Alumni Research Foundation and NSF CAREER award 1943149.
}



%% file: appendix.tex
\appendix
\begin{table}[b]
\begin{center}
\begin{tabular}{p{0.7in}p{1.2in}l}
  \toprule
\textbf{Type} & \textbf{Transformation} & \textbf{Range}\\
\midrule
\multirow{6}{0.8in}{Physical} & Rotation & $[0, 360^{\circ}]$\\
& Horizontal Flip & \{0, 1\}\\
& Vertical Flip & \{0, 1\}\\
& Relative translation & $[0, 0.7]$\\
& Relative Distance & $[1, 1.5]$\\
& Relative lighting & $[0.8, 1.2]$\\
\midrule
\multirow{2}{0.8in}{Color Error (per channel)} &
Affine additive & $[-0.2, 0.2]$\\
&
Affine multiplicative & $[0.7, 1.3]$\\

\bottomrule
\end{tabular}
\end{center}
\caption{Ranges for the transformation parameters used for generating and evaluating signals}
\label{tab:par-dis}
\end{table}

\section{Distributions of Transformations}
To make our adversarial signal effective in a physical setting, we use the EOT framework. 
We choose a distribution of transformations. The optimization produces an adversarial
example that is robust under the distribution of transformations. \tabref{tab:par-dis} describes the transformations.

\paragraph{Physical transformations.}
The relative translation involves moving the object in the image's field of view. A translation value of 0 means the object is in the center of the image, while a value of 1 means the object is at the boundary of the image. The relative distance transform involves enlarging the object to emulate a closer distance. A distance value of 1 is the same as the original image, while for the value of 1.5, the object is enlarged to 1.5 times the original size. 

\paragraph{Color correction.} Moreover, we apply a multiplicative brightening transformation to the ambient light image to account for small changes in ambient light. To account for the color correction, we used an affine transform of the form $Ax + B$, where $A$ and $B$ are real values sampled from a uniform distribution independently for each color channel.

\section{Additional Simulation Results}
For evaluating the attack in a simulated setting, we select 5 classes from the ImageNet dataset. We select 7 target classes for each source class and report the results in Table \ref{tab:simulation-full-results}. The attack generation and evaluation is the same as described previously. The attack success rate is calculated as the percentage of images classified as the target among 200 transformed images each averaged over all the possible signal offsets.  \figref{tab:simulation-results-ball}, \ref{tab:simulation-results-tbear} and \ref{tab:simulation-results-rifle} give a random sample of 4 transformed images for 3 source classes. For each source class, we give attacked images for 3 target classes.


  


\begin{table*}[t]
\centering
\begin{tabular}{p{0.7in}p{1.2in}cc}
  \toprule
\textbf{Source (confid.)} & \textbf{Affinity targets} & \multicolumn{1}{p{0.5in}}{\textbf{Attack success}} & \multicolumn{1}{p{1.1in}}{\textbf{Target confidence (StdDev)}} \\
\midrule
\multirow{3}{0.8in}{Coffee mug (83\%)} 
  & Perfume & 99\% & 82\% (13\%)\\
  & Petri dish & 98\% & 88\% (15\%)\\
  & Candle & 98\% & 85\% (18\%)\\
  & Menu & 97\% & 84\% (16\%)\\
  & Lotion & 91\% & 75\% (17\%)\\
  & Ping-pong ball & 79\% & 68\% (27\%)\\
  & Pill bottle & 23\% & 40\% (17\%)\\
  \midrule
\multirow{3}{0.8in}{Street sign (87\%)} & Monitor       & 99\% & 94\% (12\%)\\
  & Park bench & 99\% & 90\% (13\%)\\
  & Lipstick & 84\% & 78\% (20\%)\\
  & Slot machine & 48\% & 59\% (19\%)\\
  & Carousel & 41\% & 61\% (25\%)\\
  & Pool table & 34\% & 47\% (19\%)\\
  & Bubble & 26\% & 37\% (22\%)\\
  \midrule
 \multirow{3}{0.8in}{Teddy bear (93\%)} & Tennis ball    & 92\% & 88\% (19\%)\\
& Sock    & 76\% & 57\% (22\%)\\
  & Acorn    & 75\% & 72\% (25\%)\\
  & Pencil box    & 69\% & 48\% (20\%)\\
  & Comic book    & 67\% & 44\% (18\%)\\
  & Hour glass    & 64\% & 53\% (25\%)\\
  & Wooden spoon    & 62\% & 53\% (22\%)\\
  
  \midrule
\multirow{3}{0.8in}{Soccer ball (97\%)}
  & Pinwheel      & 96\% & 87\% (15\%)\\
  & Goblet      & 78\% & 55\% (17\%)\\
  & Helmet      & 66\% & 59\% (22\%)\\
  & Vase      & 44\% & 44\% (17\%)\\
  & Table lamp      & 43\% & 46\% (14\%)\\
  & Soap dispenser      & 37\% & 34\% (16\%)\\
  & Thimble      & 10\% & 15\% (02\%)\\
  \midrule
\multirow{3}{0.8in}{Rifle (96\%)} & Bow & 76\% & 64\% (24\%)\\
& Microphone              & 74\% & 63\% (22\%)\\
  & Tripod              & 65\% & 65\% (22\%)\\
  & Tool kit              & 57\% & 56\% (22\%)\\
  & Dumbbell              & 35\% & 44\% (21\%)\\
  & Binoculars              & 35\% & 40\% (18\%)\\
  & Space bar              & 17\% & 33\% (17\%)\\
  
\bottomrule
\end{tabular}
\caption{Performance of affinity targeting using our adversarial light signals
  on five classes from ImageNet. For each source
  class we note the top 7 affinity targets, their attack success rate, and
  average classifier confidence of the target class. (Average is taken
    over all offsets values for 200 randomly sampled transformations.)}
\label{tab:simulation-full-results}
\end{table*}

\newcommand{\imgszz}{3.7cm}

\begin{figure*}[p]
\begin{center}
\begin{tabular}{cccc}
  \toprule
\textbf{Original - Teddy Bear} & \textbf{Sock} & \textbf{Pencil box} & \textbf{Hour glass}\\
\midrule
\includegraphics[width=\imgszz]{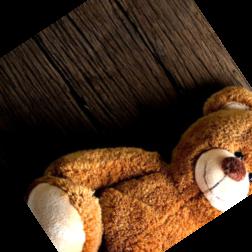} &
\includegraphics[width=\imgszz]{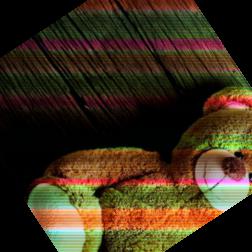} &
\includegraphics[width=\imgszz]{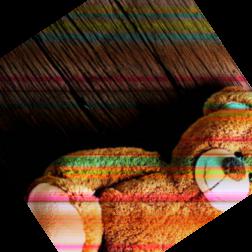} &
\includegraphics[width=\imgszz]{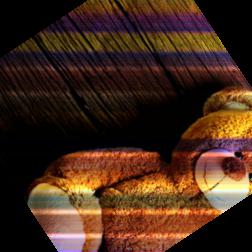}
\\
97\% & 90\% & 25\% & 20\%\\
\midrule
\includegraphics[width=\imgszz]{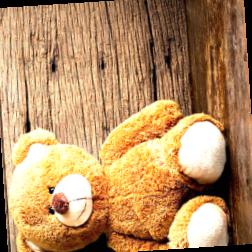} &
\includegraphics[width=\imgszz]{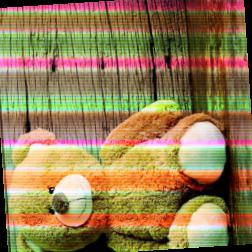} &
\includegraphics[width=\imgszz]{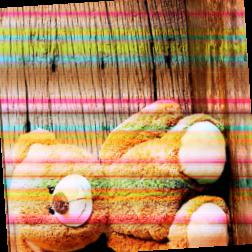} &
\includegraphics[width=\imgszz]{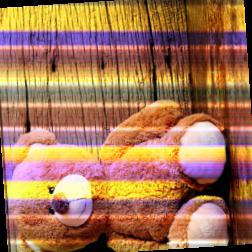}
\\
100\% & 83\% & 66\% & 61\%\\
\midrule
\includegraphics[width=\imgszz]{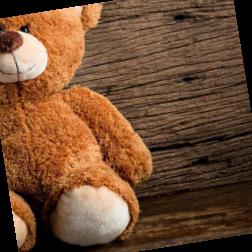} &
\includegraphics[width=\imgszz]{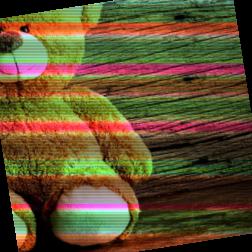} &
\includegraphics[width=\imgszz]{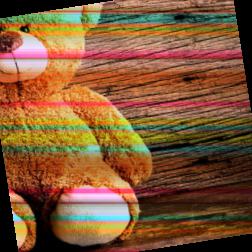} &
\includegraphics[width=\imgszz]{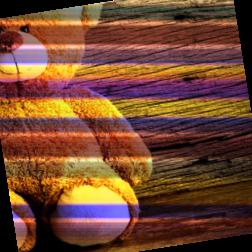}\\
100\% & 91\% & 40\% & 83\%\\
\midrule
\includegraphics[width=\imgszz]{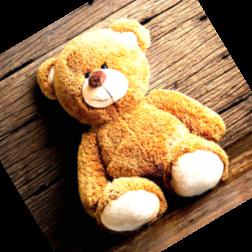} &
\includegraphics[width=\imgszz]{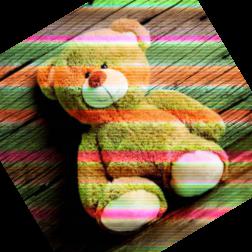} &
\includegraphics[width=\imgszz]{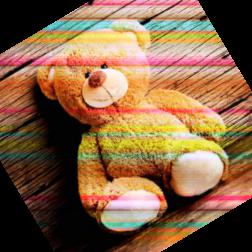} &
\includegraphics[width=\imgszz]{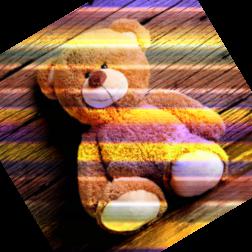} \\
100\% & 78\% & 88\% & 86\%\\
\bottomrule
\end{tabular}
\end{center}
\caption{A random sample of targeted attacks against class - Teddy Bear. The attack is robust to viewpoint, distance and small lighting changes. The numbers denote the confidence values for the respective classes. }
\label{tab:simulation-results-tbear}
\end{figure*}

\begin{figure*}[p]
\begin{center}
\begin{tabular}{cccc}
  \toprule
\textbf{Original - Soccer ball} & \textbf{Pinwheel} & \textbf{Goblet} & \textbf{Helmet}\\
\midrule
\includegraphics[width=\imgszz]{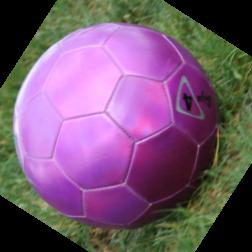} &
\includegraphics[width=\imgszz]{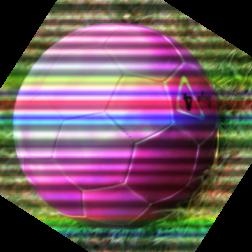} &
\includegraphics[width=\imgszz]{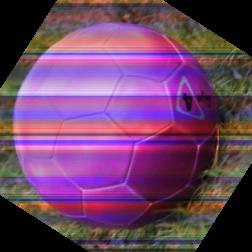} &
\includegraphics[width=\imgszz]{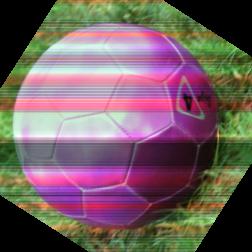}\\
100\% & 96\% & 54\% & 70\%\\
\midrule
\includegraphics[width=\imgszz]{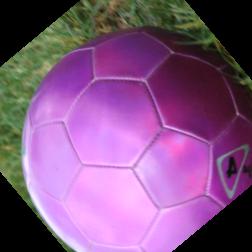} &
\includegraphics[width=\imgszz]{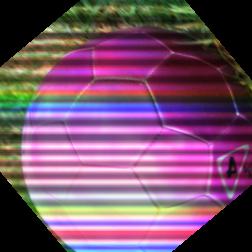} &
\includegraphics[width=\imgszz]{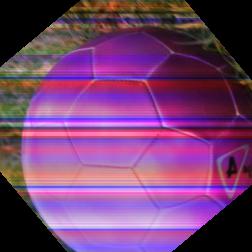} &
\includegraphics[width=\imgszz]{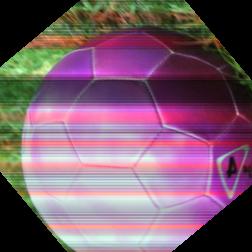}\\
98\% & 98\% & 73\% & 58\%\\
\midrule
\includegraphics[width=\imgszz]{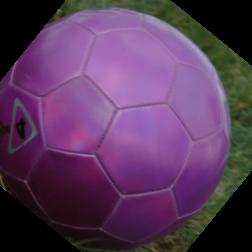} &
\includegraphics[width=\imgszz]{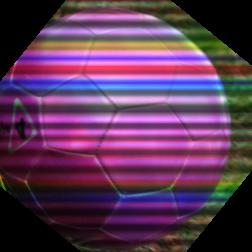} &
\includegraphics[width=\imgszz]{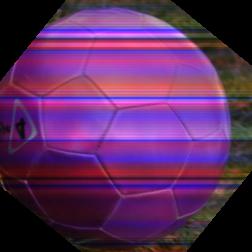} &
\includegraphics[width=\imgszz]{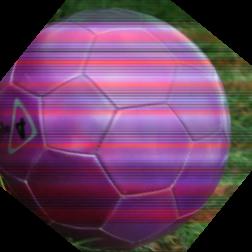}\\
90\% & 83\% & 32\% & 40\%\\
\midrule
\includegraphics[width=\imgszz]{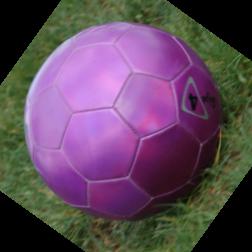} &
\includegraphics[width=\imgszz]{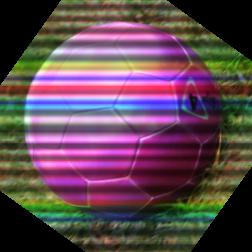} &
\includegraphics[width=\imgszz]{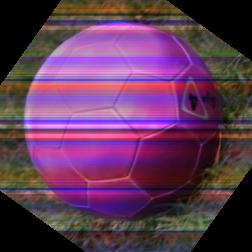} &
\includegraphics[width=\imgszz]{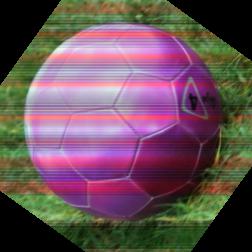}\\
99\% & 88\% & 55\% & 24\%\\
\bottomrule
\end{tabular}
\end{center}
\caption{A random sample of targeted attacks against class - Soccer ball. The attack is robust to viewpoint, distance and small lightning changes. The numbers denote the confidence values for the respective classes. }
\label{tab:simulation-results-ball}
\end{figure*}

\begin{figure*}[p]
\begin{center}
\begin{tabular}{cccc}
  \toprule
\textbf{Original - Rifle} & \textbf{Bow} & \textbf{Microphone} & \textbf{Tool kit} \\
\midrule
\includegraphics[width=\imgszz]{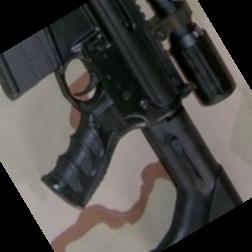} &
\includegraphics[width=\imgszz]{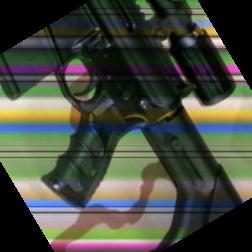} &
\includegraphics[width=\imgszz]{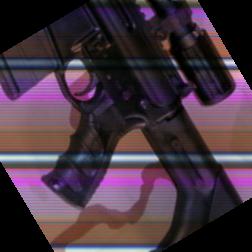} &
\includegraphics[width=\imgszz]{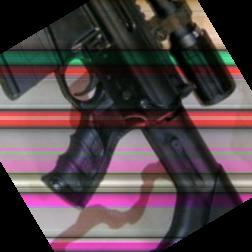}
\\
81\% & 94\% & 32\% & 70\%\\
\midrule
\includegraphics[width=\imgszz]{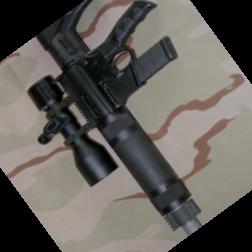} &
\includegraphics[width=\imgszz]{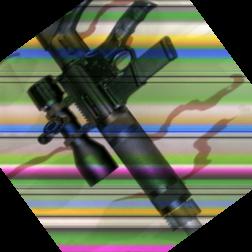} &
\includegraphics[width=\imgszz]{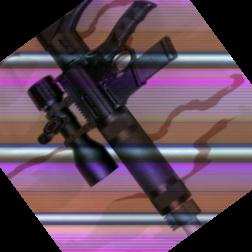} &
\includegraphics[width=\imgszz]{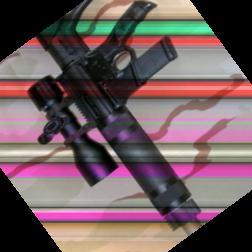}
\\
77\% & 100\% & 87\% & 50\%\\
\midrule
\includegraphics[width=\imgszz]{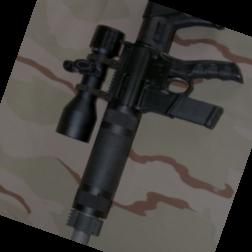} &
\includegraphics[width=\imgszz]{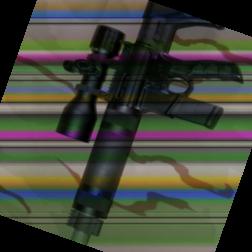} &
\includegraphics[width=\imgszz]{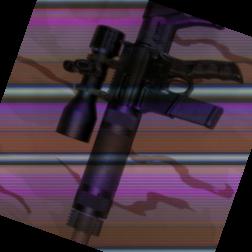} &
\includegraphics[width=\imgszz]{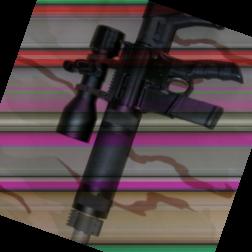}
\\
66\% & 98\% & 56\% & 72\%\\
\midrule
\includegraphics[width=\imgszz]{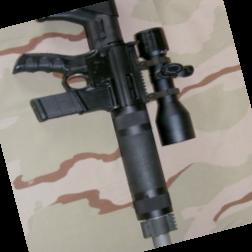} &
\includegraphics[width=\imgszz]{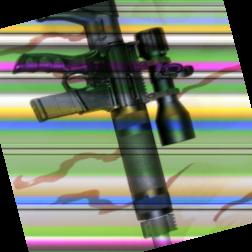} &
\includegraphics[width=\imgszz]{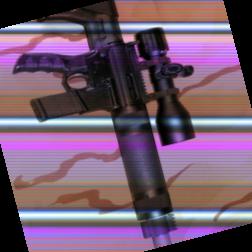} &
\includegraphics[width=\imgszz]{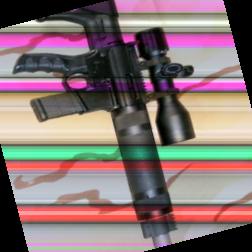}\\
65\% & 100\% & 29\% & 77\%\\
\bottomrule
\end{tabular}
\end{center}
\caption{A random sample of targeted attacks against class - Rifle. The attack is robust to viewpoint, distance and small lightning changes. The numbers denote the confidence values for the respective classes.}
\label{tab:simulation-results-rifle}
\end{figure*}

%% file: main.bbl
\begin{thebibliography}{10}\itemsep=-1pt

\bibitem{alliedvision}
Allied vision: {ALVIUM} 1800 u-1240.
\newblock
  https://www.alliedvision.com/en/products/embedded-vision-cameras/detail/Alvium

\bibitem{onsemi}
{AR023Z}: {CMOS} image sensor, 2 mp, 1/2.7".
\newblock
  https://www.onsemi.com/products/sensors/image-sensors-processors/image-sensors/ar023z.

\bibitem{albl2020rolling}
Cenek Albl, Zuzana Kukelova, Viktor Larsson, Tomas Pajdla, and Konrad
  Schindler.
\newblock From two rolling shutters to one global shutter, 2020.

\bibitem{srgbstandard}
Matthew Anderson, Ricardo Motta, S. Chandrasekar, and Michael Stokes.
\newblock Proposal for a standard default color space for the internet - srgb.
\newblock In {\em Color Imaging Conference}, 1996.

\bibitem{eot}
Anish Athalye, Logan Engstrom, Andrew Ilyas, and Kevin Kwok.
\newblock Synthesizing robust adversarial examples.
\newblock volume~80 of {\em Proceedings of Machine Learning Research}, pages
  284--293, Stockholmsmässan, Stockholm Sweden, 10--15 Jul 2018. PMLR.

\bibitem{athalye2017synthesizing}
Anish Athalye and Ilya Sutskever.
\newblock Synthesizing robust adversarial examples.
\newblock {\em arXiv preprint arXiv:1707.07397}, 2017.

\bibitem{biggio2013evasion}
Battista Biggio, Igino Corona, Davide Maiorca, Blaine Nelson, Nedim
  {\v{S}}rndi{\'c}, Pavel Laskov, Giorgio Giacinto, and Fabio Roli.
\newblock Evasion attacks against machine learning at test time.
\newblock In {\em Joint European Conference on Machine Learning and Knowledge
  Discovery in Databases}, pages 387--402. Springer, 2013.

\bibitem{bou2010controller}
Haitham Bou-Ammar, Holger Voos, and Wolfgang Ertel.
\newblock Controller design for quadrotor uavs using reinforcement learning.
\newblock In {\em Control Applications (CCA), 2010 IEEE International
  Conference on}, pages 2130--2135. IEEE, 2010.

\bibitem{bradley_synchronization_2009}
Derek Bradley, Bradley Atcheson, Ivo Ihrke, and Wolfgang Heidrich.
\newblock Synchronization and rolling shutter compensation for consumer video
  camera arrays.
\newblock In {\em {{IEEE Computer Society Conference}} on {{Computer Vision}}
  and {{Pattern Recognition Workshops}}}, pages 1--8, {Miami, FL}, June 2009.
  {IEEE}.

\bibitem{patch-attack}
Tom Brown, Dandelion Mane, Aurko Roy, Martin Abadi, and Justin Gilmer.
\newblock Adversarial patch.
\newblock 2017.

\bibitem{carlini2017towards}
Nicholas Carlini and David Wagner.
\newblock Towards evaluating the robustness of neural networks.
\newblock In {\em Security and Privacy (SP), 2017 IEEE Symposium on}, pages
  39--57. IEEE, 2017.

\bibitem{chia-kai_liang_analysis_2008}
{Chia-Kai Liang}, {Li-Wen Chang}, and H.H. Chen.
\newblock Analysis and {{Compensation}} of {{Rolling Shutter Effect}}.
\newblock {\em IEEE Transactions on Image Processing}, 17(8):1323--1330, Aug.
  2008.

\bibitem{deng2009imagenet}
Jia Deng, Wei Dong, Richard Socher, Li-Jia Li, Kai Li, and Li Fei-Fei.
\newblock Imagenet: A large-scale hierarchical image database.
\newblock In {\em Computer Vision and Pattern Recognition, 2009. CVPR 2009.
  IEEE Conference on}, pages 248--255. IEEE, 2009.

\bibitem{roadsigns17}
Kevin Eykholt, Ivan Evtimov, Earlence Fernandes, Bo Li, Amir Rahmati, Chaowei
  Xiao, Atul Prakash, Tadayoshi Kohno, and Dawn Song.
\newblock {Robust Physical-World Attacks on Deep Learning Visual
  Classification}.
\newblock In {\em Computer Vision and Pattern Recognition (CVPR)}, June 2018.

\bibitem{geiger2012we}
Andreas Geiger, Philip Lenz, and Raquel Urtasun.
\newblock Are we ready for autonomous driving? the kitti vision benchmark
  suite.
\newblock In {\em Computer Vision and Pattern Recognition (CVPR), 2012 IEEE
  Conference on}, pages 3354--3361. IEEE, 2012.

\bibitem{geyer_geometric_2005}
Christopher Geyer, Marci Meingast, and Shankar Sastry.
\newblock Geometric {{Models}} of {{Rolling}}-{{Shutter Cameras}}.
\newblock In {\em Proc. {{Omnidirectional Vision}}, {{Camera Networks}} and
  {{Non}}-Classical {{Cameras}}}, pages 12--19, 2005.

\bibitem{goodfellow2014explaining}
Ian~J Goodfellow, Jonathon Shlens, and Christian Szegedy.
\newblock Explaining and harnessing adversarial examples.
\newblock {\em arXiv preprint arXiv:1412.6572}, 2014.

\bibitem{disco}
Kensei Jo, Mohit Gupta, and Shree~K. Nayar.
\newblock Disco: Display-camera communication using rolling shutter sensors.
\newblock 35(5), July 2016.

\bibitem{kim2020object}
Namhoon Kim, Junsu Bae, Cheolhwan Kim, Soyeon Park, and Hong-Gyoo Sohn.
\newblock Object distance estimation using a single image taken from a moving
  rolling shutter camera.
\newblock {\em Sensors}, 20(14):3860, 2020.

\bibitem{kingma2014adam}
Diederik Kingma and Jimmy Ba.
\newblock Adam: A method for stochastic optimization.
\newblock {\em arXiv preprint arXiv:1412.6980}, 2014.

\bibitem{carsec}
Karl Koscher, Alexei Czeskis, Franziska Roesner, Shwetak Patel, Tadayoshi
  Kohno, Stephen Checkoway, Damon McCoy, Brian Kantor, Danny Anderson, Hovav
  Shacham, and Stefan Savage.
\newblock Experimental security analysis of a modern automobile.
\newblock In {\em Proceedings of the 2010 IEEE Symposium on Security and
  Privacy}, SP '10, page 447–462, USA, 2010. IEEE Computer Society.

\bibitem{kurakin2016adversarial}
Alexey Kurakin, Ian Goodfellow, and Samy Bengio.
\newblock Adversarial examples in the physical world.
\newblock {\em arXiv preprint arXiv:1607.02533}, 2016.

\bibitem{rolling-light}
Hui-Yu Lee, Hao-Min Lin, Yu-Lin Wei, Hsin-I Wu, Hsin-Mu Tsai, and Kate Ching-Ju
  Lin.
\newblock Rollinglight: Enabling line-of-sight light-to-camera communications.
\newblock In {\em Proceedings of the 13th Annual International Conference on
  Mobile Systems, Applications, and Services}, MobiSys '15, page 167–180, New
  York, NY, USA, 2015. Association for Computing Machinery.

\bibitem{cam-sticker}
Juncheng Li, Frank Schmidt, and Zico Kolter.
\newblock Adversarial camera stickers: A physical camera-based attack on deep
  learning systems.
\newblock volume~97 of {\em Proceedings of Machine Learning Research}, pages
  3896--3904, Long Beach, California, USA, 09--15 Jun 2019. PMLR.

\bibitem{liang2008analysis}
Chia-Kai Liang, Li-Wen Chang, and Homer~H Chen.
\newblock Analysis and compensation of rolling shutter effect.
\newblock {\em IEEE Transactions on Image Processing}, 17(8):1323--1330, 2008.

\bibitem{lillicrap2015continuous}
Timothy~P Lillicrap, Jonathan~J Hunt, Alexander Pritzel, Nicolas Heess, Tom
  Erez, Yuval Tassa, David Silver, and Daan Wierstra.
\newblock Continuous control with deep reinforcement learning.
\newblock {\em arXiv preprint arXiv:1509.02971}, 2015.

\bibitem{rswhitepaper}
J. Linkemann and B. Weber.
\newblock Global shutter, rolling shutter—functionality and characteristics
  of two exposure methods (shutter variants).
\newblock {\em White Paper}, 2014.

\bibitem{madry2017towards}
Aleksander Madry, Aleksandar Makelov, Ludwig Schmidt, Dimitris Tsipras, and
  Adrian Vladu.
\newblock Towards deep learning models resistant to adversarial attacks.
\newblock {\em arXiv preprint arXiv:1706.06083}, 2017.

\bibitem{moosavi2015deepfool}
Seyed-Mohsen Moosavi-Dezfooli, Alhussein Fawzi, and Pascal Frossard.
\newblock Deepfool: a simple and accurate method to fool deep neural networks.
\newblock {\em arXiv preprint arXiv:1511.04599}, 2015.

\bibitem{mostegel2016uav}
Christian Mostegel, Markus Rumpler, Friedrich Fraundorfer, and Horst Bischof.
\newblock Uav-based autonomous image acquisition with multi-view stereo quality
  assurance by confidence prediction.
\newblock In {\em Proceedings of the IEEE Conference on Computer Vision and
  Pattern Recognition Workshops}, pages 1--10, 2016.

\bibitem{openpilot}
OpenPilot.
\newblock {OpenPilot on the Comma Two}.
\newblock \url{https://github.com/commaai/openpilot}, 2020.

\bibitem{hololens}
TOMMY PALLADINO.
\newblock Hololens 2, all the specs — these are the technical details driving
  microsoft's next foray into augmented reality.
\newblock
  https://hololens.reality.news/news/hololens-2-all-specs-these-are-technical-details-driving-microsofts-next-foray-into-augmented-reality-0194141/,
  2019.

\bibitem{papernot2016limitations}
Nicolas Papernot, Patrick McDaniel, Somesh Jha, Matt Fredrikson, Z~Berkay
  Celik, and Ananthram Swami.
\newblock The limitations of deep learning in adversarial settings.
\newblock In {\em Security and Privacy (EuroS\&P), 2016 IEEE European Symposium
  on}, pages 372--387. IEEE, 2016.

\bibitem{sharif2016accessorize}
Mahmood Sharif, Sruti Bhagavatula, Lujo Bauer, and Michael~K Reiter.
\newblock Accessorize to a crime: Real and stealthy attacks on state-of-the-art
  face recognition.
\newblock In {\em Proceedings of the 2016 ACM SIGSAC Conference on Computer and
  Communications Security}, pages 1528--1540. ACM, 2016.

\bibitem{flicker}
M. {Sheinin}, Y.~Y. {Schechner}, and K.~N. {Kutulakos}.
\newblock Rolling shutter imaging on the electric grid.
\newblock In {\em 2018 IEEE International Conference on Computational
  Photography (ICCP)}, pages 1--12, 2018.

\bibitem{szegedy2014intriguing}
Christian Szegedy, Wojciech Zaremba, Ilya Sutskever, Joan Bruna, Dumitru Erhan,
  Ian Goodfellow, and Rob Fergus.
\newblock Intriguing properties of neural networks.
\newblock In {\em International Conference on Learning Representations}, 2014.

\bibitem{adv-t-shirt}
Kaidi Xu, Gaoyuan Zhang, Sijia Liu, Quanfu Fan, Mengshu Sun, Hongge Chen,
  Pin-Yu Chen, Yanzhi Wang, and Xue Lin.
\newblock Adversarial t-shirt! evading person detectors in a physical world.
\newblock In Andrea Vedaldi, Horst Bischof, Thomas Brox, and Jan-Michael Frahm,
  editors, {\em Computer Vision -- ECCV 2020}, pages 665--681, Cham, 2020.
  Springer International Publishing.

\bibitem{zhang2015towards}
Fangyi Zhang, J{\"u}rgen Leitner, Michael Milford, Ben Upcroft, and Peter
  Corke.
\newblock Towards vision-based deep reinforcement learning for robotic motion
  control.
\newblock {\em arXiv preprint arXiv:1511.03791}, 2015.

\bibitem{lishield}
Shilin Zhu, Chi Zhang, and Xinyu Zhang.
\newblock Automating visual privacy protection using a smart led.
\newblock MobiCom '17, page 329–342, New York, NY, USA, 2017. Association for
  Computing Machinery.

\end{thebibliography}
